\documentclass[runningheads]{llncs}

\usepackage[utf8]{inputenc}
\usepackage{graphicx}
\usepackage{comment}
\usepackage{amsmath,amssymb} %
\usepackage{color}

\usepackage{tabularx} %

\usepackage{acronym}
\usepackage{siunitx}
\usepackage{bm}
\usepackage{xspace}
\usepackage{microtype}

\usepackage{booktabs}

\usepackage{subcaption}
\usepackage{xcolor}
\usepackage{siunitx}
\usepackage{multirow}

\usepackage{tikz}
\usepackage{pgf}
\usepackage{pgfplots}
\usepgfplotslibrary{groupplots}
\pgfplotsset{compat=newest}
\usetikzlibrary{shadows,spy,decorations,plotmarks,matrix,mindmap,%
arrows,arrows.meta,patterns,fit,calc,positioning,shapes}

\usepackage[hyphens]{url}

\usepackage[pagebackref=true,breaklinks=true,colorlinks,bookmarks=false,citecolor=green!80!black,linkcolor=red!80!black,urlcolor=blue]{hyperref}

\usepackage{cleveref}

\crefname{section}{Sec.}{Sections}
\crefname{figure}{Fig.}{Figures}
\crefname{table}{Tab.}{Tables}
\crefname{equation}{Eq.}{Equations}
\crefname{appsec}{Appendix}{Appendices}

\usepackage[inline]{enumitem}
\setlist*[enumerate]{label=(\arabic*)}

\makeatletter
\DeclareRobustCommand\onedot{\futurelet\@let@token\@onedot}
\def\@onedot{\ifx\@let@token.\else.\null\fi\xspace}
\makeatother
\newcommand{\etal}[1]{#1~\textit{et~al\onedot}}

\newcommand{\eg}{e.\,g.,\xspace}

\newcommand{\cf}{cf\onedot}
\newcommand{\ie}{i.\,e.,\xspace}

\newcommand{\turing}{Turing\xspace}
\newcommand{\style}{Style\xspace}

\newacro{cnn}[CNN]{Convolutional Neural Network}
\newacro{cvrnn}[CVRNN]{Conditional VRNN}
\newacro{dl}[DL]{Deep Learning}
\newacro{gan}[GAN]{Generative Adversarial Network}
\newacro{htr}[HTR]{Handwritten Text Recognition}
\newacro{lstm}[LSTM]{Long Short-Term Memory}
\newacro{rnn}[RNN]{Recurrent Neural Network}
\newacro{stcnn}[STCNN]{Stochastic Temporal CNN}
\newacro{vrnn}[VRNN]{Variational RNN}
\newacro{oov}[OOV]{out-of-vocabulary}

\newcommand{\imgheight}{0.023\textheight}
\newcolumntype{Y}{>{\centering\arraybackslash}X}

\definecolor{uncontext}{HTML}{0180d2} 
\definecolor{context}{HTML}{157f2d} 

\newenvironment{customlegend}[1][]{%
	\begingroup
	\csname pgfplots@init@cleared@structures\endcsname
	\pgfplotsset{#1}%
}{%
	\csname pgfplots@createlegend\endcsname
\endgroup
}%
\def\addlegendimage{\csname pgfplots@addlegendimage\endcsname}
\newcommand{\addlegendimageintext}[1]{%
	\tikz {
		\begin{customlegend}[
				legend entries={\empty},
				legend style={
					draw=none,
					inner sep=0pt,
					column sep=0pt,
				nodes={inner sep=0pt}}]
				\addlegendimage{#1}
			\end{customlegend}
		}%
	}

\graphicspath{{img/}}

\begin{document}

\pagestyle{headings}
\mainmatter

\title{Spatio-Temporal Handwriting Imitation}

\renewcommand*{\vec}[1]{\mathbf{#1}}

\title{Spatio-Temporal Handwriting Imitation}

\titlerunning{Spatio-Temporal Handwriting Imitation}
\author{Martin Mayr\orcidID{0000-0002-3706-285X} \and
Martin Stumpf\orcidID{0000-0002-2278-8500} \and \\
Anguelos Nicolaou\orcidID{0000-0003-3818-8718} \and
Mathias Seuret\orcidID{0000-0001-9153-1031} \and
Andreas Maier\orcidID{0000-0002-9550-5284} \and
Vincent Christlein\orcidID{0000-0003-0455-3799}}
\authorrunning{M.\ Mayr et al.}
\institute{Pattern Recognition Lab, Friedrich–Alexander-Universität Erlangen–Nürnberg
\email{\{firstname.lastname\}@fau.de}\\
\url{https://lme.tf.fau.de/}}

\maketitle

\begin{abstract}
Most people think that their handwriting is unique and cannot be imitated by machines, especially not using completely new content. 
Current cursive handwriting synthesis is visually limited or needs user interaction.
We show that subdividing the process into smaller subtasks makes it possible to imitate someone's handwriting with a high chance to be visually indistinguishable for humans.
Therefore, a given handwritten sample will be used as the target style. 
This sample is transferred to an online sequence. Then, a method for online handwriting synthesis is used to produce a new realistic-looking text primed with the online input sequence. This new text is rendered and style-adapted to the input pen. We show the effectiveness of the pipeline by generating in- and out-of-vocabulary handwritten samples that are validated in a comprehensive user study. Additionally, we show that also a typical writer identification system can partially be fooled by the created fake handwritings.

\end{abstract}

\begin{figure}[h]
    \centering
			 \begin{tikzpicture}
				 \node[draw,rounded corners](i1){
					 \includegraphics[height=0.6cm]{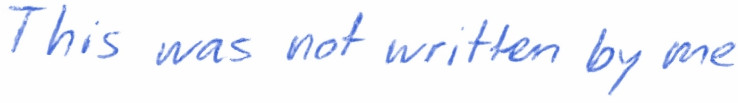}
				 };
				 \node[right=1.5cm of i1,draw,rounded corners](i2){
    			\includegraphics[height=0.6cm]{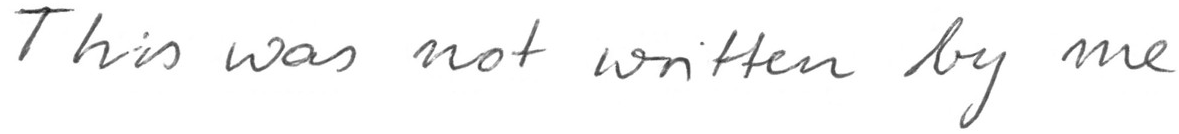}
				};
				\draw[<->,thick,>=stealth](i1)--node[midway,text
				width=4em,align=center]{real or fake?\footnotemark}(i2);
			 \end{tikzpicture} 
		\label{fig:teaser}
\end{figure}

\section{Introduction}
\footnotetext{Both samples are synthesized.}

Handwriting is still a substantial part of communication, note making, and authentication. 
Generating text in your handwriting without the need to actually take a pen in your hand can be beneficial, not only because we live in the age of digitization but also when the act of writing is physically impaired due to injuries or diseases.
Handwriting synthesis could also enable to send handwritten messages in a much more personal way than using standard handwriting fonts, \eg for gift messages when sending presents. 
Personal handwriting could also be useful in virtual reality games, where parts could be written in a famous handwriting or in the player's own handwriting in order to identify more strongly with the avatar.
Similarly, it could be used for augmented reality or in movies. Why not adapting someone's handwriting style when translating a note from one language into another one?
Since the most deep learning methods need large datasets for training, another use case could be the improvement of automatic \ac{htr}. 
In fact, the simulation of single handwritten words is already used for data augmentation during the training of \ac{htr} methods~\cite{Alonso19}.

With the help of the method we propose, it is possible to generate handwritten text in a personal handwriting style from only one \emph{single} handwritten line of text of the person to be imitated. 
This method would also enable the creation of realistic-looking forgeries.
To detect these forgeries, new and robust writer identification/verification methods may need to be developed.
To train such systems, our method allows forensic researchers to generate and imitate handwriting without the need of skillful forgers.
\begin{figure}[t]
    \centering
		\tikzset{
			     block/.style={rectangle, draw, text width=1.7cm,
						                    text centered, minimum
															height=1cm},
					arrow/.style={shorten >=3pt, shorten <=3pt,->,thick}
					}
		\begin{tikzpicture}[>=stealth]
				\node[block](in){
					\includegraphics[width=1.4cm]{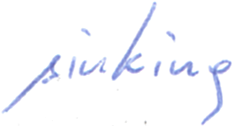}
				};
				\node[block,right=2.8cm of in](skel){
					\includegraphics[width=1.4cm]{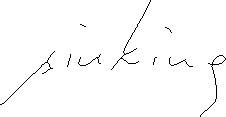}
				};
				\node[block,right=2.8cm of skel](online){
					\includegraphics[width=1.4cm]{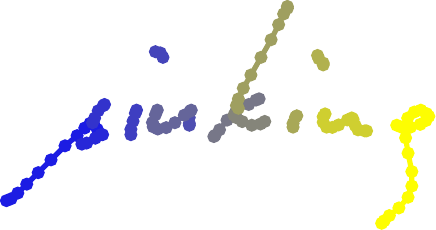}
				};
				\node[block,below=1.5cm of online](wi-style){
					\includegraphics[width=1.4cm]{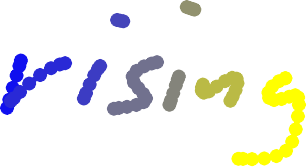}
				};
				\node[block,below=1.5cm of skel](render){
					\includegraphics[width=1.4cm]{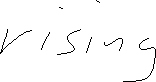}
				};
				\node[block,below=1.5cm of in](out){
					\includegraphics[width=1.4cm]{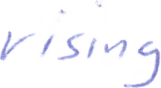}
				};
				\draw[arrow]($(in.east)-(0cm,0.1cm)$) -| ($(out.east) + (2.45cm,0cm)$);
				
				\draw[arrow](in)--node[above]{Skeletonization}(skel);
				\draw[arrow](skel)--node[above]{Online} node[below]{Approximation}(online);
				\draw[arrow](online)--node[anchor=north east]{Writer Style} node[anchor=north west]{Transfer\phantom{y}}(wi-style);
				\draw[arrow](wi-style)--node[above]{Skeleton} node[below]{Rendering}(render);
				\draw[arrow](render)--node[above]{Image Style} node[below]{Transfer}(out);
				
				\draw[arrow]($(online.south)-(-1.3cm,0.6cm)$) --
				node[anchor=south]{\phantom{MM}``rising''} ($(online.south) - (-0.cm,0.6cm)$);
				
			\end{tikzpicture}
    \caption{Offline-to-Offline Handwriting Style Transfer Pipeline}
    \label{fig:pipeline}
\end{figure}
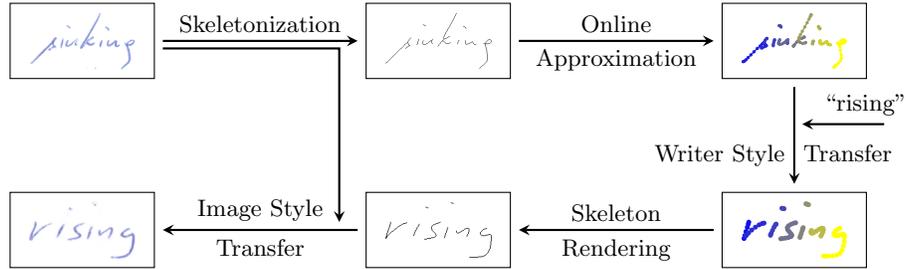

In contrast to \emph{offline} data, \ie scanned-in handwriting, \emph{online} handwriting stores for each data point not only its position but also the temporal information, representing the actual movement of a pen. 
This can for example be recorded with special pens or pads. 
In our pipeline, the offline data is transferred to the temporal domain, see \cref{fig:pipeline} top branch.
Then, we make use of an existing online handwriting synthesis method~\cite{graves}, which takes the temporal information and the target text as input. 
Finally, the resulting online data is rendered again in the style of the writer to be imitated, \cf \cref{fig:pipeline} bottom branch.

The main contributions of this work are as follows: \begin{enumerate*}
\item creation of a full pipeline for the synthesis of artificial Western cursive handwriting, recreating both the visual and the writer-specific style;
\item development of a novel conversion method to approximate online handwriting from offline data to be able to utilize an existing online-based style transfer method;
\item the adaptation of conditional \acp{gan} to compute a robust handwriting skeletonization and a visual style transfer to adapt to the used pen. For the former, we introduce an iterative knowledge transfer to make the offline skeletons more similar to the online training data. For the latter, we modified the well-known pix2pix framework~\cite{Isola17} to incorporate the writing style information. 
\item Finally, our method is thoroughly evaluated in two ways. 
First, by means of a writer identification method to quantitatively assess that the writing style is preserved and second, by a user study to evaluate to what extent humans can be fooled by the generated handwriting. 
\end{enumerate*}

\section{Related Work}
~\\\noindent\textbf{Online Handwriting Synthesis.}\quad
To produce convincing handwriting, it needs to reproduce the given content exactly, while keeping the style consistent, but not constant. 
Real human handwriting will repeat the same content almost identically, but still with some variance. 
This requires a solid long term memory, combined with some guidance from the required content.
In the seminal work by Alex Graves~\cite{graves} this is achieved by the use of a \ac{rnn} with \ac{lstm} cells, which enables the network to make predictions in context of the previous pen positions. 
Each hidden layer uses skip connections to give the network control over the learning flow. 
Additionally, an early form of attention mechanism was developed that decides which part of the content the network focuses on.
A problem of Graves' method~\cite{graves} is that it tends to change the writing during the sequence generation. 
This can be overcome by the use of \acp{vrnn} to generate consistent handwriting~\cite{Chung15}. 
Another method~\cite{Aksam18} builds on the idea of predicting single pen positions.
Instead of relying on the network's internal memory to store style, the goal was to explicitly extract style and content from the input. 
This is achieved by utilizing a \ac{cvrnn} to split the input into two separate latent variables, representing style and content.
A drawback of this method is the need to split the input strokes into words and characters in form of begin- and end-of-character (EOC) tokens, which typically cannot be automatically determined from offline handwriting.
In contrast to Graves' attention mechanism, it uses the generated EOC tokens to switch letters.
This could remove some of the predictive capability of the network, since it can
only foresee which letter follows when the next letter is already about to be written.
In a subsequent work~\cite{Aksan19}, the \ac{cvrnn} was replaced by 
\acp{stcnn} showing more consistent handwriting generation. 

~\\\noindent\textbf{Offline Handwriting Synthesis.}\quad
There are approaches that use printed text style synthesis for augmentation~\cite{Gomez19} or text stylization for artistic typography creation~\cite{Yang19}. 
Similarly, cursive handwriting synthesis~\cite{Alonso19} can be generated using a \ac{gan} combined with an auxiliary text recognition network. 
While the augmentation of \ac{htr} improves the recognition accuracy, the generated words cannot imitate a specific handwriting. 
Another approach~\cite{Balreira17} synthesis handwriting from public fonts by finding the best character matches in public handwritten fonts.
The results are convincing but still far away from the actual user's style.

The closest work to ours is the work by \etal{Haines}~\cite{Haines2016}. 
To the best of our knowledge, this is the only other method using offline cursive handwritings as input. 
They produce convincing output by creating glyph models from segmented labeled ligatures and individual glyphs. The method has two main short-comings. 
First, the selection of the glyphs and ligatures involves human assistance.
Second, only letters present in the handwriting to be imitated can be reproduced.

Another work~\cite{Lian18} converted offline data to another modality and then rendered and adapted the style. 
Therefore, glyphs need to be segmented and matched to characters.
From these matched glyphs, strokes are extracted by registering them to a trajectory database,
and sampled with regular points.
In contrast, we propose a method based on maximum acceleration that uses more points for curved strokes. 
The user's style in~\cite{Lian18} is learned by a feed forward neural network,
which is added during the rendering process.  
While no human interaction is needed to generate handwritten Chinese fonts with a large amount of characters, the ``characters should be written separately without touching each other in a given order and consistently in size and style.''~\cite{Lian18}.
This issue is targeted by \etal{Nakamuar}~\cite{Nakamura18} who generate Chinese characters for samples with an incomplete character set by choosing the closest learned character distribution.

In contrast, our method works fully automatic for handwriting without any user interaction given a single line for priming.
It is able to render consistent full lines instead of single glyphs/words as in other works.
The method is able to produce \ac{oov} letters and words as long as the trained model has seen some instances during training, but not necessarily from the writer to be imitated.

\section{Offline-to-Offline Handwriting Style Transfer Pipeline}

We decided to split the offline to offline handwriting style transfer system into several subtasks.
While background, pen and writer style are static problems that could be solved with \acp{cnn}, the text content has a structural component and therefore makes some temporal generation, \eg in form of an \ac{rnn} preferable.
Each of the subtasks is trainable on its own, allowing human prior knowledge to guide the process, and to evaluate  single steps separately. 
Furthermore, not all these tasks require a neural network if ther is an algorithmic solution.
For a writer imitation, we need to apply a style transfer on two levels: 
\begin{enumerate*}
    \item the arrangement of writing strokes and
    \item the pen style, \ie thickness and color distribution.
\end{enumerate*}
\Cref{fig:pipeline} gives an overview over our 
pipeline.\footnote{Code and
	models available below \url{https://github.com/M4rt1nM4yr/spatio-temporal_handwriting_imitation}}
First, a skeleton is computed from the input sequence, which is used as the writing style to imitate (commonly several words long).
The skeleton is converted to online data with a novel sampling process that puts emphasis on the curved structure of handwriting. 
Afterwards, we make use of the writing generation method of Graves~\cite{graves}, which creates new online text in a given writing style. 
This handwriting sequence is rendered as a skeleton and transferred to the visual appearance of the priming sequence using additional data created through the initial skeletonization process.
Each step of the pipeline is described in the following subsections. 

\begin{figure}[t]
    \centering
    \subcaptionbox{\label{fig:cycle_fail}}[.35\linewidth]{
			 \begin{tikzpicture}
				 \node[draw](i1){
					 \includegraphics[height=0.7cm]{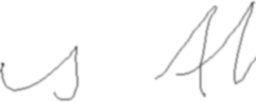}
				 };
				 \node[draw,right=0.05cm of i1](i2){
					 \includegraphics[height=0.7cm]{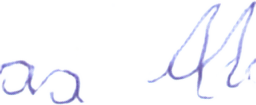}
				 };
		 \end{tikzpicture}
		 }%
		 \,
    \subcaptionbox{\label{fig:sharp_vs_nn}}[.63\linewidth]{
			\centering
			 \begin{tikzpicture}
				 \node[draw](i1){
    			\includegraphics[height=0.7cm]{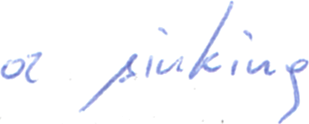}
				};
				 \node[draw,right=0.05cm of i1](i2){
    			\includegraphics[height=0.7cm]{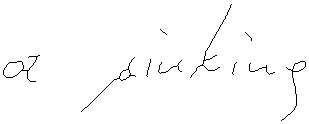}
				};
				 \node[draw,right=0.05cm of i2](i3){
			    \includegraphics[height=0.7cm]{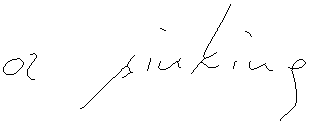}
				};
			 \end{tikzpicture}  
    }
		\caption{\subref{fig:cycle_fail} shows a skeleton rendered from the
		IAM-Online dataset (left) and the estimated offline handwriting by the CycleGAN
		(right). \subref{fig:sharp_vs_nn} shows an offline example excerpt (left),
		skeleton output of a basic method~\cite{Zhang84} (middle), and through our
	skeletonization network (right).}
		\label{fig:skeleton_examples}
\end{figure}
\subsection{Skeletonization}\label{sec:skeletonization}

In this stage, we convert images of real handwriting to their corresponding skeletons which are then subsequently mapped to the temporal domain and fed into the generative network. %
There are sophisticated learning-free~\cite{Bai20} and \ac{dl}~\cite{Shen17}-based methods, which learn a mapping between natural objects and skeletons. 
We face the challenge that individual datasets exist of both real handwriting images and skeletons (online data). 
However, to the best of our knowledge, there is no dataset that annotates a mapping between those two.

\begin{figure}[t]
    \centering
		\tikzstyle{block}=[draw,thick,rounded corners,text
						width=6ex,align=center]
		\sffamily				

		\begin{tikzpicture}[>=stealth]
			\node[](d1){Domain A:};
			\node[below=1.5cm of d1](d2){Domain B:};

			\node[block,right=0.7cm of d1](r1){Real};
			\node[block,below=1.5cm of r1](s1){Synth};
			\draw[->,thick] (r1.south west) to [out=225,in=135]
			node[midway,left,ellipse,minimum width=1cm,minimum height=1.5em,align=center,inner sep=1]{Naive}
		  node[pos=0.2,right,circle,draw,inner sep=1pt,xshift=3pt]{1}
			(s1.north	west) ; 
			\draw[->,thick,dotted] (s1.north east) to [out=45,in=315]
			node[pos=0.2,left,circle,draw,inner	sep=1pt,xshift=-3pt,solid]{2}
			(r1.south east){};

			\node[block,right=2cm of r1](r2){Synth};
			\node[block,below=1.5cm of r2](s2){Real};
			\draw[->,thick] (s2.north west) to [out=135,in=225]
		  node[pos=0.2,right,circle,draw,inner sep=1pt,xshift=3pt]{3}
			(r2.south west){};
			\draw[->,thick,dotted] (r2.south east) to [out=315,in=45]
			node[pos=0.2,left,circle,draw,inner	sep=1pt,xshift=-3pt,solid]{4}
			(s2.north east){};
			\node[] (nn1) at ($(r1)!0.5!(s2)$) {\includegraphics[width=1cm]{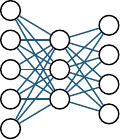}};
			\node[below=0cm of nn1,inner sep=0pt,outer sep=0pt](t1){NN1};
			
			\node[block,right=2cm of r2](r3){Real};
			\node[block,below=1.5cm of r3](s3){Synth};
			\draw[->,thick] (r3.south west) to [out=225,in=135]
		  node[pos=0.2,right,circle,draw,inner sep=1pt,xshift=3pt]{5}
			(s3.north west){};

			\node[] (nn1) at ($(r2)!0.5!(s3)$) {\includegraphics[width=1cm]{img/nnet.pdf}};
			\node[below=0cm of nn1,inner sep=0pt,outer sep=0pt](t1){NN2};
		  \node[fit=(d1)(d2)(s3)(r3),inner sep=0pt,outer sep=0pt](graph){};

			\node[right=1.2cm of r3.north east,anchor=north west](gen){\scriptsize{Generate}};
      \node[left=0.4cm of gen](c1){};
      \node[right=0.4cm of c1](c2){};
      \draw[->,thick](c1)--(c2);

			\node[below=0.1cm of gen.south west,anchor=north west](train){\scriptsize{Train}};
			\node[left=0.4cm of train](c11){};
      \node[right=0.4cm of c11](c21){};
      \draw[->,dotted,thick](c11)--(c21);
      \node[fit=(c1)(gen)(c11)(train),draw,outer sep=0pt,inner sep=0pt](legend){};

		\end{tikzpicture}
		\caption{\textbf{Iterative knowledge transfer from naive algorithms.}
(1)~A naive algorithm transfers real data from domain A to fake data of domain B. 
(2)~Using this fake data, a neural network (NN1) is trained, (3)~which is then used to convert real data from domain B to domain A. (4)~With this new data the final NN2 is trained, which (5)~can then be used 
for producing synthetic data. 
}
    \label{fig:method_iterative_knowledge_transfer}
\end{figure}

Given an offline handwritten sample, the challenge is to produce a skeleton similar enough to the online data used in the generative network. 
Specifically, we use the CVL dataset~\cite{cvl} as the source for real offline
data and the IAM-Online dataset~\cite{iam-online} for real skeleton data, where the latter is also used for training the generative network. 
A natural choice would be the use of CycleGAN~\cite{Zhu17}, which enforces cycle consistency, \ie the output of a transfer from source to target and back is similar to the source again (and vice versa).
A shortcoming of CycleGAN is that it has to guess the transfer function. 
This could lead to several problems because it is not guaranteed that the resulting mapping will still satisfy spatial consistency. 
It could freely add/remove strokes, as long as the result is \ac{gan}-consistent and contains enough information to perform an inverse mapping, see \cref{fig:cycle_fail} for an example.
Conversely, there are basic skeletonization algorithms~\cite{Zhang84,Lee94}. 
These could be used to guide the training to incorporate prior knowledge about the mapping, \cf~\cref{fig:sharp_vs_nn}. 
Therefore, we propose \emph{iterative knowledge transfer} to extract the knowledge of one of those algorithms and transfer it to a neural network.
The proposed method is not limited to our specific use case.
It is rather a general method to transfer knowledge from an existing mapping function to a neural network while improving and generalizing it along the way.
It requires a naive mapping function and two non-paired datasets for which we would like to achieve a mapping and consists of the following steps, as illustrated in \cref{fig:method_iterative_knowledge_transfer}.

\begin{enumerate}
\item Generation of synthetic skeletons from the real source dataset (CVL) using the naive mapping. This dataset is expected to be erroneous. 
\item Training of an inverse mapping (NN1) based on the synthetic dataset, enforcing generalization. 
The network capacity needs to be sized correctly, so that generalization happens without picking up on the errors of the naive implementation.
\item Generation of synthetic CVL data from the real skeletons (IAM Online) using the trained mapping. 
\item This new dataset is used to train the desired mapping from offline handwriting to their corresponding skeletons (NN2). 
\item The final model is expected to produce realistic output data, because the target of the training procedures was the generation of real target data.
\end{enumerate}
It is assumed that in steps 2 and 4 the synthetic data is almost similar to the real data, otherwise the training procedure would not produce reasonable results.

\subsection{Online Sequence Approximation}

The purpose of this stage of the pipeline is to take skeleton images as input and convert them to an approximate online representation.
As skeletons do not contain temporal annotations, this step will require a heuristic approach to synthesize realistic online data analytically. 
This stage consists of two steps:
\begin{enumerate*}
\item conversion of the bitmap representation to strokes, and
\item temporal resampling and ordering.
\end{enumerate*}

\subsubsection{Conversion to Strokes.}\quad
First, the bitmap data is converted to a graph. This is done by connecting the
neighboring pixels because we assume that strokes are connected lines in the
skeleton. 
Our target is to create strokes without cycles, but the generated graph contains pixel clusters.
These pixel clusters are defined as triangles having at least one common edge. Graph cycles that do not consist of triangles are assumed to be style characteristics of the writer, like drawing the dot on the `i' as a tiny circle.
The cluster artifacts are solved by replacing all cluster nodes that are not connected to outside nodes with a single node at the mean position of the group.
The final step for creating strokes is to remove intersections and cycles. 
This implies that the nodes of every line segment have either one or two neighbors.
To produce a consistent behaviour the cycles are split at the upmost position, similarly to human writing.

\subsubsection{Resampling.}
\begin{figure}[t]
	\centering
	\tikz{
		\node[draw](i1){%
			\footnotesize{(a)}\phantomsubcaption\label{fig:res_non}
			\includegraphics[width=0.13\textwidth]{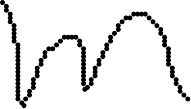}
		};
		\node[draw, right=1.5cm of i1](i2){%
			\footnotesize{(b)}\phantomsubcaption\label{fig:res_constant}
			\includegraphics[width=0.13\textwidth]{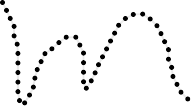}
		};
		\node[draw,right=1.5cm of i2]{%
		\footnotesize{(c)}\phantomsubcaption\label{fig:res_maxacc}
			\includegraphics[width=0.13\textwidth]{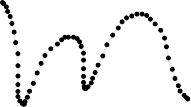}
		};
	}
	\caption{Visualisation of different resampling methods. \subref{fig:res_non} no
	resampling, \subref{fig:res_constant} constant velocity resampling,
	\subref{fig:res_maxacc} proposed maximum acceleration resampling.}
  \label{fig:resamplingComparison}
\end{figure}

\quad
To obtain online handwriting, we still need to incorporate temporal information.
The simplest way would be to sample at constant time steps.
However, the training of the subsequent online network becomes very difficult because very small time steps are needed. 
We were unable to generate convincing results, presumably because the sequence is too far away from a real handwriting sequence.
Therefore, we propose \emph{maximum acceleration resampling} on the computed lines to imitate the dynamics of human writing.
We constrain the velocity to be zero at both extremities of a line, increase it on straight parts,
and decrease it on curved parts. 
This has the advantage that the network focuses on important parts of the
handwriting.
The difference between constant resampling and the proposed maximum acceleration
resampling can be seen in \cref{fig:resamplingComparison}.
In detail, the algorithm consists of the following steps:
\begin{enumerate*}
\item Resampling of the curve to sufficiently small intervals.
\item Creating a reachability graph between nodes to prevent cutting corners.
\item Analyzing acceleratability, \ie the acceleration required between two
	nodes. This will push the problem into a 4D space: ($x$,$y$,$v_x$,$v_y$).
\item Searching shortest path using directed Dijkstra.
\end{enumerate*}

\paragraph{Constant Pre-sampling.} The Dijkstra search~\cite{dijkstra59} does not
include the possibility to cut lines into pieces. It can only select a set of
optimal nodes from existing nodes. Consequently, we need to make sure that
the existing nodes are spaced appropriately by resampling them into small
constant intervals. 
We empirically set the distance to $1/3$ of the maximum acceleration value.

\paragraph{Reachability Graph.} We have to avoid cutting corners, such that during
curved sections the sampling rate is higher.
Therefore, we create a graph, encoded in a boolean matrix of size $N\times N$
(where $N$ is the number of nodes in the pre-sampled graph), which stores the
pairwise reachability between all nodes.
We define points $\vec{p}_i$ and $\vec{p}_{i+n}$ to be reachable, when $\max_{j=i..i+n} d(\vec{p}_j) < t$ with $d(\vec{p}_j)$ being the distance of the point $\vec{p}_j$ to the line
between $\vec{p}_i$ and $\vec{p}_{i+n}$ and $t$ being a given threshold.
We found the threshold parameter to be optimal at $3$ times the
node distance of the pre-sampling step.

\paragraph{Acceleratability.} So far, our nodes are two dimensional: $\vec{p} = (x,
y)$. We now enhance them with two more dimensions: the incoming velocities
$\vec{v} = (v_x, v_y)$. The incoming velocity $\vec{v}_i$ of node $\vec{p}_i$
from node $\vec{p}_{i-n}$ is computed by 
$ \vec{v}_i = \vec{p}_i - \vec{p}_{i-n} $.
Thus, only one 4D node exists for each preceding 2D node, with multiple possible edges from different velocities of that preceding node.
We now create a 4D acceleratability graph based on the 2D reachability graph
that connects all 4D nodes $(\vec{p}_i, \vec{v}_i)$ and $(\vec{p}_j, \vec{v}_j)$
that fulfill the following `acceleratability' criterion:
\begin{equation}\label{eq:acceleratability}
\vec{v}_j = \vec{p}_j - \vec{p}_i \;\; \land \;\; ||\vec{v}_j - \vec{v}_i|| < a
\end{equation}
with $a$ being the maximum acceleration hyperparameter.
As we never go back, we have a directed graph, and thus it contains all possible pen trajectories that create the given curve.

\paragraph{Shortest Path Search.}
The set of possible paths is quite large, therefore we use a Dijkstra shortest path search with some optimizations specific to the 4D case.
Since we have a directed graph, we will always start at one end of the stroke and
move towards the other one. So, we can step through the
curve, 2D node by 2D node, and compute all optimal paths to that node for
all possible velocities at that node. The number of possible velocities
is limited and is equivalent to the number of incoming edges to that node.
Computing an optimal path to a given position $\vec{p}$ and velocity $\vec{v}$ can be done as follows. First, we get the position of the previous node:
$ \vec{p}_{prev} = \vec{p} - \vec{v} $.
We now take all the possible velocities $\vec{v}_{prev}$ at position
$\vec{p}_{prev}$ that can reach $\vec{p}$, based on the acceleratability
criterion, \cf \cref{eq:acceleratability}.
The shortest path $l$ to $(\vec{p}, \vec{v})$ is then:
\begin{equation}
l(\vec{p}, \vec{v}) = \min_{\vec{v}_{prev}} l(\vec{p}_{prev},\vec{v}_{prev}) + 1
\end{equation}

We start the entire algorithm at one end of the curve, which we define as the
starting point. Further, we set $\vec{p} = \vec{p}_{start}$, for which the only
valid velocity is $\vec{v}_{start} = 0$ and $l(\vec{p}_{start}, \vec{v}_{start})
= 0$. We then iterate through the entire curve until we reach $\vec{p}_{end}$.
As we defined both start and end velocities to be zero, we then look at
$l(\vec{p}_{end}, \vec{v}_{end})$ with $\vec{v}_{end} = 0$ and backtrack to get
the optimal path.

\paragraph{Ordering.}
Finally, the points are ordered 
by computing the mean of every stroke and ordering them from left to right to sort the list of strokes. 
In real human writing there are cases where this is not true, but in
this way we produce a consistent behaviour, which is necessary for the further stages in the pipeline.

\subsection{Writer Style Transfer}

The produced online sequence is used to prime an online writing synthesis
algorithm.
We employ Graves' algorithm~\cite{graves} who showed that \ac{lstm} cells are capable
of generating complex structures with long-term contextual dependencies.
To be able to predict both the content and the style at the same time, 
the content of the text is added to the network as a side input to one of the
intermediate layers. 
The network does not see the entire content sequence at once, instead, a form of
attention is used. 
To achieve this, the mixture density output from intermediate layers
decide which part of the content gets delivered to the network.
Note that we are not forced to use Gaves' method but can use any online
handwriting generation approach. 

\subsection{Image Style Transfer}\label{sec:img_style_transfer}
The produced new sequence of online data is transferred back to offline
handwriting by means of drawing lines between the points of the online data. 
The last stage of the pipeline is to produce realistic offline handwriting by
imitating the ink and style of the input image given the new skeleton of the
produced online handwriting. 
Therefore, we modify the pix2pix~\cite{Isola17} architecture to output the
correct style.

\begin{figure}[t]
    \centering
    \includegraphics[width=0.6\textwidth]{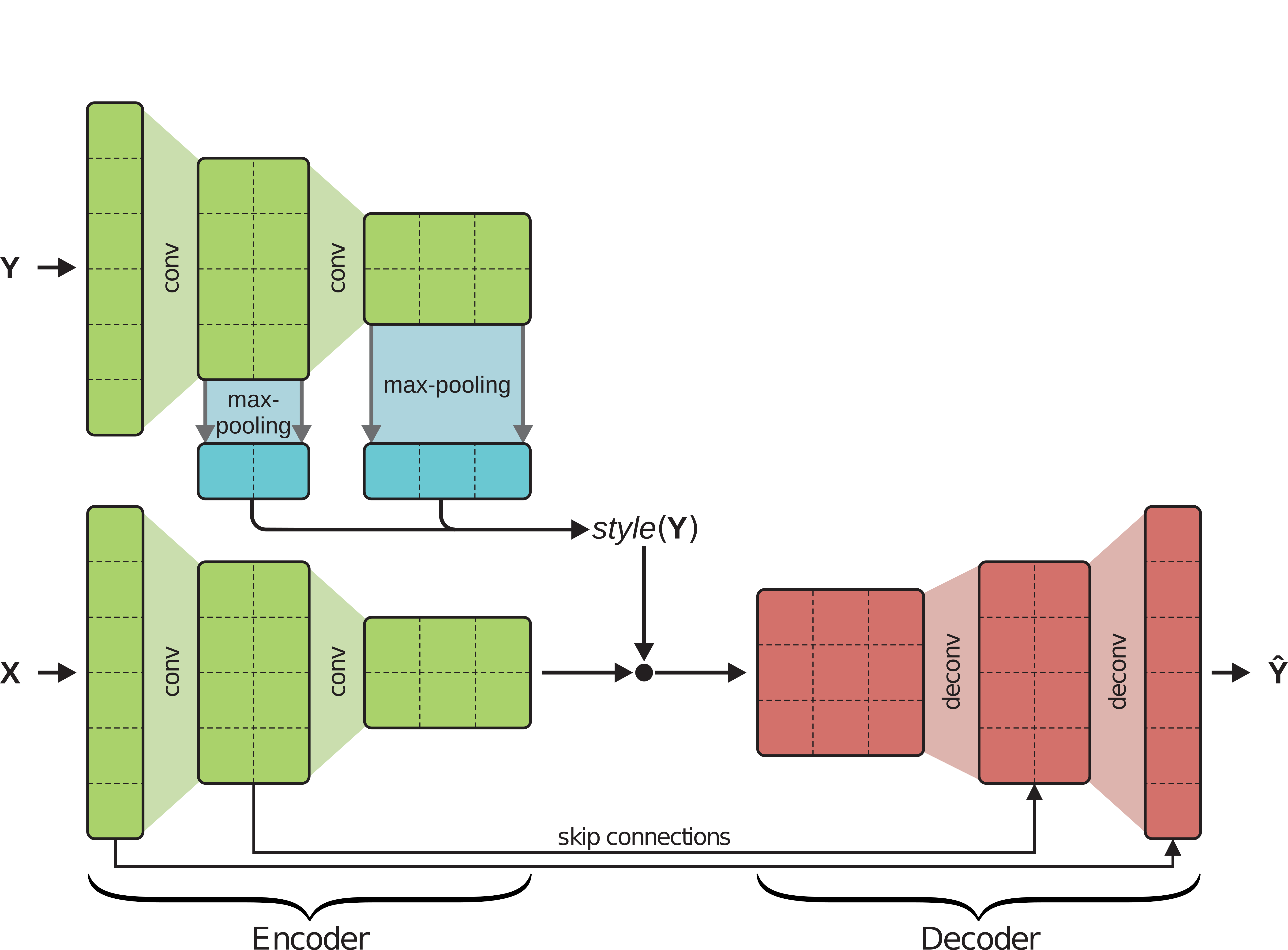}
		\caption{Modified pix2pix generator network for conditional style transfer.
			The green network depicts the encoder of the pix2pix generator network.
			The max-pooling extracts the style information of all activation maps of style image $Y$.
			The style information gets concatenated with the output of the encoder for newly rendered image $X$ used in the decoder to generate the output $\hat{Y}$.}
		\label{fig:method_style_extraction}
\end{figure}

The network consists of an encoder and a decoder network. 
The first step in creating a style transfer network is to extract the style
information from our input image $Y$. We use the encoder part of the
pix2pix network with the style image as input. 
It is important that the style only contains global information,
otherwise the discriminator \emph{D} could start to discriminate by content instead.
Hence, we take the max-pooled outputs of all activation maps as style information.
Then, we feed the style information into the pix2pix generator network by
concatenating it with the innermost layer of the network, as seen in
\cref{fig:method_style_extraction}. To keep the size-agnostic property of the
network, we repeat the style information along the two spatial axes to match
the size of the innermost layer.

To include the style extraction network in the training process, we add the
output of the style extraction network to the input of \emph{D}. 
Incorporating this into the objective of pix2pix \cite{Isola17} results in
\begin{equation}
	\begin{aligned}
		\mathcal{L}_{\text{cGAN}(G,D)} &= \mathbb{E}_{X,Y} [\log
			D(X,Y,\text{style}(Y))]  \\
			&+ \mathbb{E}_{X,Y}[\log(1-D(X,G(X,\text{style}(Y)),\text{style}(Y)))] \\
			&+ \lambda\mathbb{E}_{X,Y}[||Y-G(X,\text{style}(Y))||_1]
			\;.
	\end{aligned}
\end{equation}
Feeding the style information to \emph{D} has the effect that \emph{D} can now differentiate between real and generated images by
comparing their style output. This forces the generator \emph{G} to
generate images that include the style taken from the extraction network
to produce a style consistent output.
We encounter that the model sometimes produces
wrong pen colors while the stroke width and all other characteristics are
transferred correctly. 
Therefore, we find the most important principal component of the style image and transfer it to the output image.

\section{Evaluation}
We conduct a large-scale user study to assess how well a human could
be fooled by our method. 
Furthermore, we utilize writer identification methods to have a non-biased
measurement of how well the results are imitating the original writer.

\subsection{Datasets}

The following datasets are used to train the models, perform the experiments and user studies.

\emph{IAM On-Line Handwriting Dataset.} The IAM On-Line Handwriting Data\-base~\cite{iam-online} consists of handwritten
samples recorded on a whiteboard. The dataset consists of 221 writers with
\num{13049} isolated and labeled text lines in total containing \num{86272}
word instances of a \num{11059} large word dictionary.

\emph{CVL Dataset.}
The CVL Database~\cite{cvl} consists of offline handwriting data providing
annotated text line and word images. We used the official training set consisting of 27 writers contributing 7 texts each. 

\emph{Out-of-vocabulary Dataset.}
For the \ac{oov} words, we 
used the 2016 English and German Wikipedia corpora~\cite{Goldhahn12},
containing both \num{100000} words.
Words containing less than four characters and words already part of the CVL
training set are removed from these dictionaries.
For the user study, the number of German words is reduced such that the ratio of
English and German words is equal to the ones of the CVL dataset.

\subsection{Implementation Details}\label{sec:implementation}

For the skeletonization, we make use of the proposed iterative knowledge transfer algorithm due to the nonexistent dataset for our problem description, see \cref{sec:skeletonization}. We use the pix2pix framework~\cite{Isola17} as the mapping network, trained with augmentations, such as added noise, resizing, cropping and color jitter.

For the pen style transfer, we use another pix2pix model for which we employ an asymmetric version of the U-Net~\cite{Ronneberger15}. It is shortened to 4 layers (original 8) and reduced number of filters: [16, 32, 64, 64] for the encoder, [32, 64, 128, 128] for the style extractor, and [192, 256, 128, 64] for the decoder.
The goal of this network is to synthesize a real-looking handwriting image given a skeleton, which was produced earlier through the pix2pix skeletonization, and a random sample of the CVL dataset as style input, \cf \cref{sec:img_style_transfer}.
Both models are trained with standard pix2pix~\cite{Isola17} training parameters.
We use ADAM with a learning rate of $0.0002$, momentum parameters $\beta_1 = 0.5$, $\beta_2 = 0.999$, 
and $\lambda_{L_1} = 100$ for weighting the $L_1$ distance between $Y$ and $\hat{Y}$.

The \ac{rnn} model of the online handwriting style transfer is trained using the IAM On-Line Handwriting Database. The
model consists of 400 LSTM cells.
It is trained with RMSProp with $\beta_1 = 0.9$, a batch size of 32 and
a learning rate of 0.0001 and gradient clipping in $[-10,10]$. 

\subsection{Qualitative results}
\begin{figure}[t]
    \centering
		\begin{tabular}{c>{\centering}p{3.4cm}>{\centering}p{3.4cm}>{\centering}p{3.4cm}}
         \toprule
         \multirow{2}{*}{(a)\phantomsubcaption\label{fig:quali_a}} & \multicolumn{3}{c}{\includegraphics[height=\imgheight]{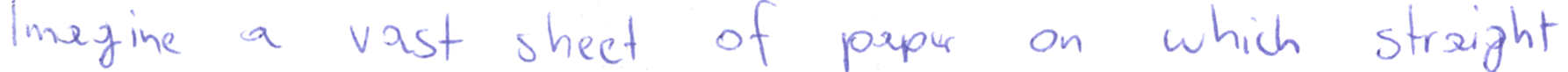}} \\ 
         & \includegraphics[height=\imgheight]{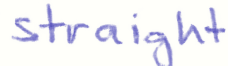} & \includegraphics[height=\imgheight]{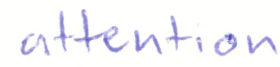} &
         \includegraphics[height=\imgheight]{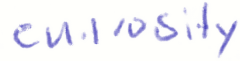}
					\tabularnewline
        \midrule
        \multirow{2}{*}{(b)\phantomsubcaption\label{fig:quali_b}}
         & \multicolumn{3}{c}{\includegraphics[height=\imgheight]{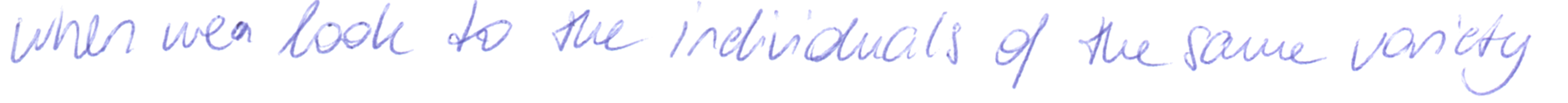}} \\ %
        & \includegraphics[height=\imgheight]{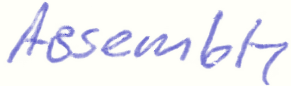} & \includegraphics[height=\imgheight]{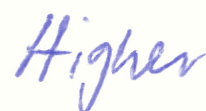} &
          \includegraphics[height=\imgheight]{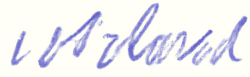}
					\tabularnewline
        \midrule
        (c)\phantomsubcaption\label{fig:quali_c}
        & 
        \multicolumn{3}{c}{\quad\;\includegraphics[height=\imgheight]{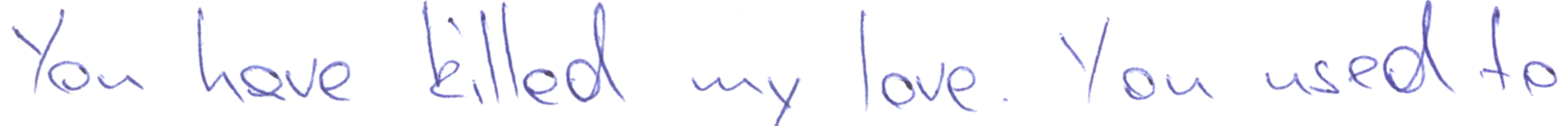}
        \includegraphics[height=\imgheight]{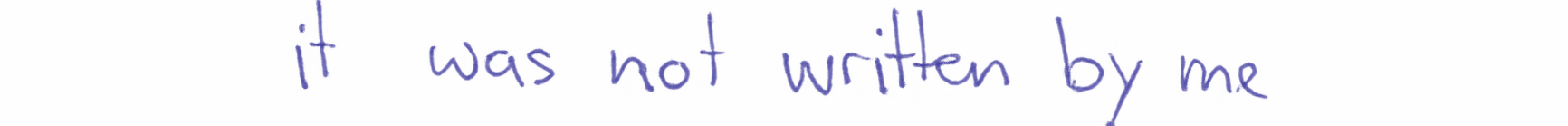}
        }
        \tabularnewline
        \midrule
         (d)\phantomsubcaption\label{fig:quali_d}
         & 
         \multicolumn{3}{c}{%
         \includegraphics[height=\imgheight]{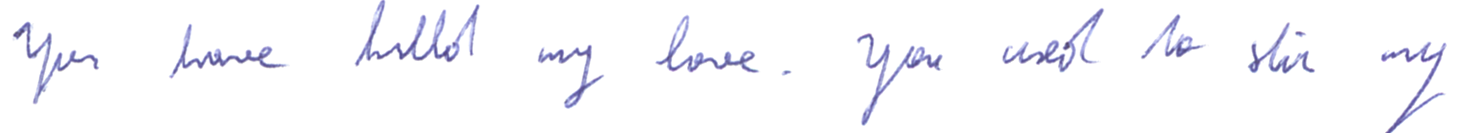}%
        \quad\quad\quad\quad
        \includegraphics[height=\imgheight]{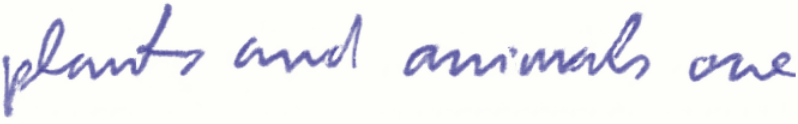}
        }
        \tabularnewline
        \midrule
         (e)\phantomsubcaption\label{fig:quali_e}
         & 
         \multicolumn{3}{c}{%
         \includegraphics[height=\imgheight]{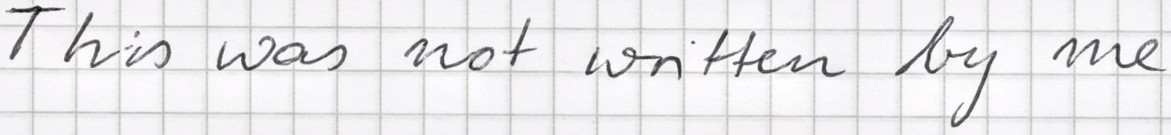}%
        \quad\quad\quad\quad
        \includegraphics[height=\imgheight]{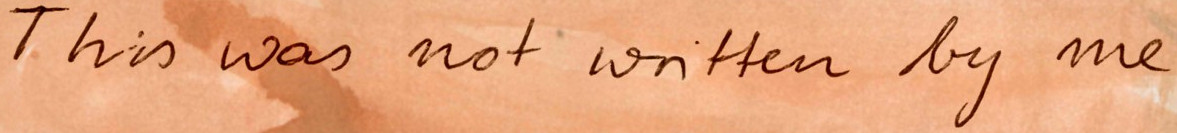}
        }
					\tabularnewline
         \bottomrule
			 \end{tabular}
    \caption{\subref{fig:quali_a} and \subref{fig:quali_b} show three generated words each - two good results and one bad one with the given style line on top. 
    \subref{fig:quali_c} and \subref{fig:quali_d} show two generated lines (right) with two style lines (left). \subref{fig:quali_e} shows synthetic outputs with added backgrounds.} 
    \label{fig:example_words}
\end{figure}
\Cref{fig:example_words} shows four distinctive outputs. 
\cref{fig:quali_a} and \cref{fig:quali_b} show three synthesized words, depicting
two good results and one containing typical pipeline failures with the genuine anchor (=style) line at top. \cref{fig:quali_c} and \cref{fig:quali_d} show two common outputs of the network with the anchor line on the left. 
For example, the addition of superfluous lines, incorrect splitting and
positioning of lines, detail removal, such as missing \emph{t}-, \emph{f}- lines
or \emph{i}-dots, or skipped letters at the beginning of the message.  
We notice that the less legible a writer's handwriting is, the harder it is to
imitate the writing style, see for example writer \cref{fig:quali_b,fig:quali_d}.
We also showcase generated lines with added backgrounds in \cref{fig:quali_e}. 

\subsection{User Study}

We conducted a user study\footnote{https://forms.gle/MGCPk5UkxnR23FqT9} to estimate the degree to which humans can distinguish real from synthesized data.
Random samples from the CVL dataset serve as real data, and the synthesized data is generated with words from the CVL and \ac{oov} corpora while line images from CVL are randomly chosen as style inputs.
Two sub-studies were performed.
In the first one~(\addlegendimageintext{draw=black,fill=uncontext,area legend}), humans are shown a sample and asked if it was
``written by a human'' or ``created by a machine''.
This is a kind of Turing test asking to differentiate between human
and synthesized handwriting.
In the second one (\addlegendimageintext{draw=black,fill=context,area legend}), 
we wanted to evaluate the ability to generate a person-specific handwriting. 
Therefore, we made an experimental setup consisting of a line written by a specific writer and two words. The user had to choose the word which was written by the same writer as the given line.
Every subject had to answer 96 questions without time limit: 32 for the
\emph{\turing} task and 64 for the \emph{\style} task. 
For each task, a background representing notebook paper was artificially
introduced~\cite{perez2003poisson,seuret2015gradient} in exactly
\SI{50}{\percent} of the queries in order to investigate the impact of such a
distraction.
In total, 59 people participated in our user study with different knowledge
background:
\emph{Humanities} (H), \ie paleographers, book scientists, etc.; people working
in \emph{Computer Vision} (CV) or image processing; general \emph{Computer
Science} (CS); and \emph{Other} (O). 
Most users needed between 8 and 24 minutes to answer the 96 questions.
\begin{figure}[t] 
	\centering

\subcaptionbox{Confusion matrix\label{tab:confusion}}
	{
	\centering
	\begin{tabular}{lccr}
	\toprule
	&  \multicolumn{2}{c}{Predicted}\\
	\cmidrule{2-3}
	& Real & Fake\\
	\midrule
		Real & 33.9 & 16.1 & R: 67.8\\
		Synth & 25.2 & 24.8 & SP: 49.6\\
		    & P: 57.4 & NPV: 60.6 & ACC: 58.7\\ %
	\bottomrule
	\\ %
\end{tabular}
}		
\subcaptionbox{Ablation Study\label{fig:by_field}}[0.5\textwidth]
{
	\centering

\begin{tikzpicture}
\begin{axis}[%
	width=0.5\textwidth,
	height=3.8cm,
	ymin=10, ymax=100,
	ylabel near ticks,
	xlabel near ticks,
	label style={font=\footnotesize},
	ymajorgrids=true,
	enlarge x limits=0.13,
	major x tick style=transparent,
	tick label style={font=\tiny},
	ybar=0, %
	ylabel={Detection rate},
		bar width=5pt,
	xticklabels={H, CV, CS, O, IV, OV, S, W},
	xtick={0,1,2,3,4,5,6,7},
	legend entries={\turing, \style},
	legend style={font=\tiny,
		/tikz/every even column/.append	style={column sep=0.5cm},
		at={(0.5,1.1)},anchor=south
	},
	legend columns=-1,
	legend cell align=left
]
				
\addlegendimage{draw=black,fill=uncontext,area legend}
\addlegendimage{draw=black,fill=context,area legend}

\addplot+[fill=uncontext,draw opacity=0] table 
	[x expr=\coordindex, y index=0] {figures/results_uncontext_all.txt};
\addplot+[fill=context,draw opacity=0] table 
	[x expr=\coordindex, y index=0] {figures/results_context_all.txt};
\end{axis}
\end{tikzpicture}
}

\caption{%
	\subref{tab:confusion} displays the results of the \turing task.
	\subref{fig:by_field} shows the results of different groups (H:
	humanities scholars, CV: computer vision/image analysis background, CS:
	computer scientist (general), O: other) and different ablation studies (OV:
	out-of-vocabulary words, IV: in-vocabulary words, S: synthetic background, W:
white background
}
\label{fig:user_study}
\end{figure}
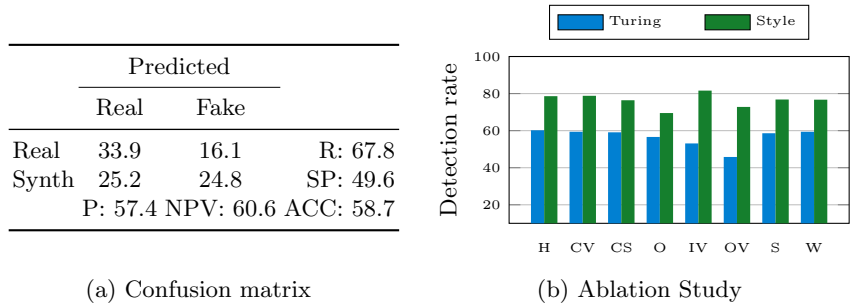

The accuracy of the \turing task is \SI{58.7}{\percent}, shown in \cref{tab:confusion}.
The accuracy of the \style task is \SI{76.8}{\percent}.
Further, in the \turing task recall (R) is higher than specificity (SP), i.e. given a real sample the decisions of the users are better.
Precision (P) and negative predictive value (NPV) are similar. %
\Cref{fig:by_field} shows that for the \turing task, the average results per user category range from \SIrange{56.6}{60.2}{\percent} correctly recognized samples. 
Experts in the humanities are slightly more accurate than others.
There is very little difference between working in the field of computer vision
or image analysis and other computer scientists (\SI{59.4}{\percent} vs.\ \SI{59.1}{\percent}).
For the \style task, the average results per user category range from \SIrange{69.5}{78.6}{\percent}.
This time, however, people with computer vision background reach an accuracy
similar to the one of scholars in the humanities (\SI{78.8}{\percent} and
\SI{78.6}{\percent}, respectively).
The gap to computer scientists and others is larger with a difference
of \SI{2.4}{\percent}, and \SI{9.3}{\percent}, respectively.
It seems that samples generated from the training set (IV) are easier to distinguish than words coming from pages of an unknown corpus (OV).
This might come from the usage of a different, more complicated lexical field.
There is no accuracy difference when using samples with synthetic
background (S) in comparison to real, white background~(W).

\subsection{Writer Identification with Synthesized Offline Handwriting}
We tested samples of our proposed method against an automatic learning-free writer identification approach~\cite{nicolaou2015sparse}.
We made a joint retrieval database having our synthetic paragraphs and a paragraph of the CVL training set from all 27 writers of the dataset, \ie a total of 54 samples.
For every query sample, we removed the real sample of the writer in question and
retrieved the most similar ones from the 53 database samples.
This procedure effectively benchmarks our forged samples vs.\ the most similar writer to the real one, which could also be considered a forgery system.
We used the specific experimental pipeline to obtain a strong baseline for forgeries and in order to asses the importance our skeletonization method trained with the proposed iterative knowledge transfer vs.\ a naive skeletonization.

\begin{table}[t]
    \caption{Evaluation of generated vs.\ nearest human writer.}
    \label{tab:srs_wi}
    \centering
    \begin{tabular}{lccc@{\hspace{1em}}cc}
\toprule
Query & \multicolumn{2}{c}{DB} & Skeleton & mAP\,\% & Acc.\,\%  \\
\midrule
Real & Real+Fake & (OV) & Naive   & 31.94   & 18.52  \\
Real & Real+Fake & (IV) & Naive   & 26.18   & \phantom{0}7.41   \\
Real & Real+Fake & (OV) & pix2pix & 29.66   & 14.82  \\
Real & Real+Fake & (IV) & pix2pix & 37.13   & 25.92  \\
\bottomrule
    \end{tabular}
\end{table}
\Cref{tab:srs_wi} (last row) shows that our method outperforms the nearest forgery on  \SI{25.9}{\percent} of the queries when our forgery is created with text from the same corpus.
When the samples are generated with out-of-corpus words, performance drops
significantly (row three).
It can also be seen that substituting our proposed skeletonization for the naive
skeletonization (row 1 and 2) has also a large impact on system performance.
As a sanity check, we tested the retrieval of real writers from real writers, naive synthetic from naive synthetic, and pix2pix synthetic from pix2pix synthetic. 
In all cases, we achieved \SI{100}{\percent} performance.
It should be pointed out that automatic methods operate on the page/paragraph level while human users were challenged at the word level. Any comparison between humans and automatic writer identification methods must not be implied from this experiment.

\section{Discussion and Limitations}
\begin{figure}[t]
	\begin{tikzpicture}
		\begin{scope}[spy using outlines={rectangle,lens={scale=3},height=1cm,width=2cm, connect spies, ultra thick}]
		\node[](orig){%
			\includegraphics[width=0.2\textwidth]{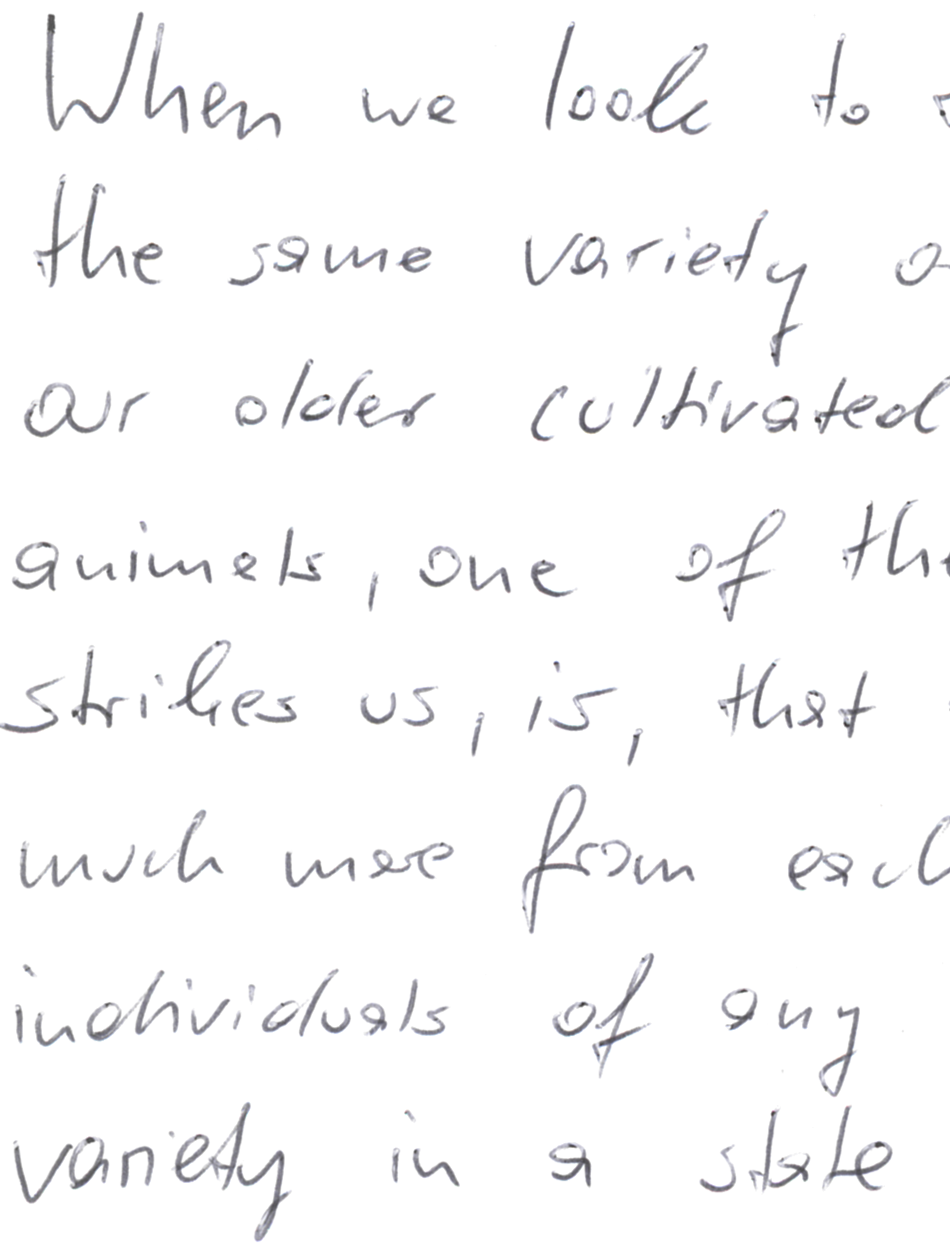}
		};

		\node[right=\textwidth of orig.west,anchor=east](fake){%
			\includegraphics[width=0.2\textwidth]{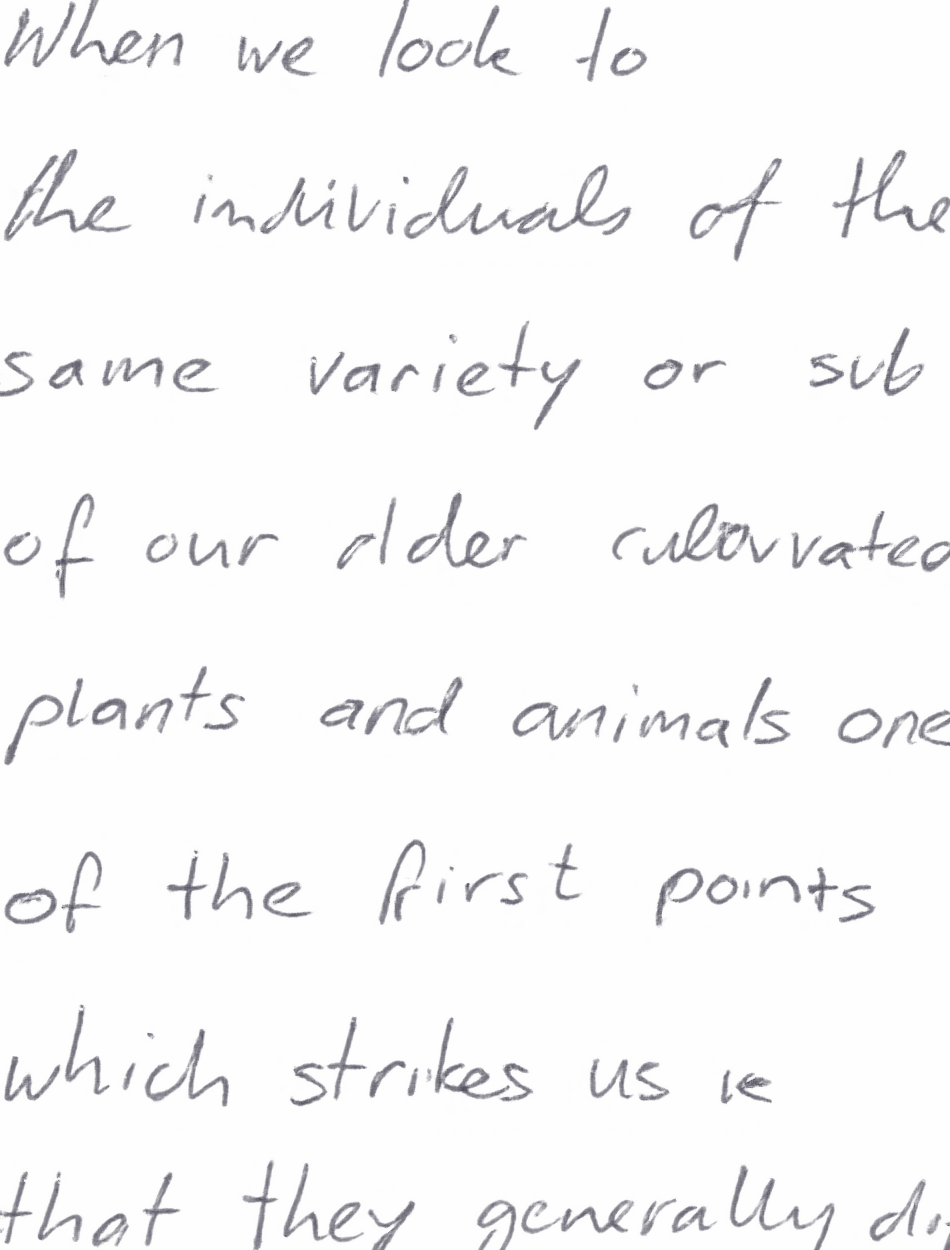}
		};

\spy[orange,thick] on (0.4cm,1.4cm) in node (s1) [right=1cm of orig.north
east,anchor=north west] ;
\spy[orange,thick] on ($(fake)+(-0.08cm,1.5cm)$) in node (s2) [left=1cm of
fake.north west,anchor=north east];
 	
\spy[orange,thick] on (0.82cm,0.19cm) in node (s3) [below=0.1cm of s1];
\spy[orange,thick] on ($(fake.west)+(0.38cm,-0.7cm)$) in node (s4) [below=0.1cm of
s2];

\spy[cyan,thick,lens={scale=10}] on ($(orig.south east)+(-0.37cm,0.35cm)$) in
node [below=0.4cm of s3] (s5);
\spy[cyan,thick,lens={scale=10}] on ($(fake.south west)+(0.44cm,0.19cm)$) in
node [below=0.4cm of s4] (s6);

\end{scope}

\node[below=0cm of orig](capa){%
	\footnotesize{(a)}\phantomsubcaption\label{fig:macro_orig}
};
\node[anchor=north] (capb) at ($(s3.south)!0.5!(s4.south)$){%
	\footnotesize{(b)}\phantomsubcaption\label{fig:stroke}
};
\node[] (capc) at (capb |- capa){%
	\footnotesize{(c)}\phantomsubcaption\label{fig:micro}
};
\node[below=0cm of fake](capd){%
	\footnotesize{(d)}\phantomsubcaption\label{fig:macro_fake}
};	
  \end{tikzpicture}
	\caption{Qualitative result of one whole paragraph: \subref{fig:macro_orig} original, \subref{fig:stroke} stroke level details, \subref{fig:micro} pen level
	details, \subref{fig:macro_fake} synthesized. }
  \label{fig:quali_level}
\end{figure}
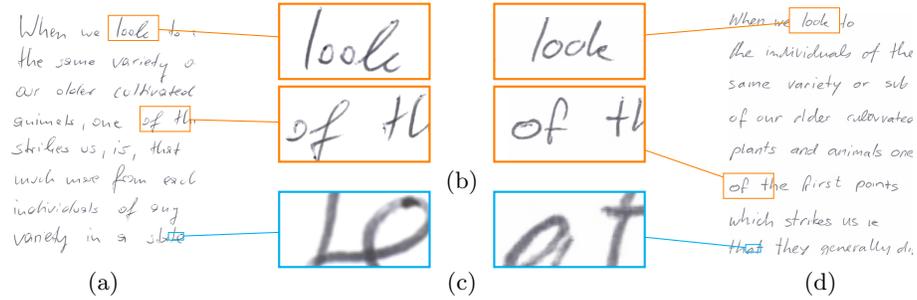
At general sight, the synthesized lines of \cref{fig:macro_fake} look very similar to
the original writing shown in \cref{fig:macro_orig}. 
But a closer look reveals flaws in the writing style. 
The orange boxes in \cref{fig:stroke} show the stroke level difference. While
there are only few minor differences in the word \emph{look} (first row), there are some major style differences in the letter \emph{f} (second row). 
On the micro level, \ie the pen style, the imitation shows in general convincing
results, \cf \cref{fig:micro}.

At the current stage, there are a few limitations. 
First of all, every step of the pipeline is dependant of the previous actions. However, we encountered that the last pipeline step is capable of fixing errors of earlier stages.
Graves model~\cite{graves} needs a minimum amount of input composed of images and
transcriptions to generate strokes with a new text sequence robustly.
This part has also difficulties to imitate writers with bad handwriting because the writing style model has problems with matching the letters of the transcription with the corresponding letters in the input image.
Our training data does not include punctuation marks, therefore the trained models are only capable of using standard characters.
In practice, scaling is also an issue, because our pipeline was mainly trained with CVL data. This can be easily solved with augmentation in upcoming versions.

\section{Conclusion}
In this work, we proposed a fully automatic method to imitate full lines of offline handwriting using spatial-temporal style transfer. 
The pipeline is capable of producing letters and words that are not in the vocabulary of the writer's samples and therefore could be applied to many use cases. 
We show that generated results just from a single line show auspicious results, often indistinguishable from real handwriting.
In future work, we will evaluate different alternatives to the chosen algorithmic steps.
We believe that especially an alternative online synthesis method can improve the system's outcome.
In the moment the pipeline is capable of changing the background in a very realistic manner~\cite{seuret2015gradient}. 
We plan to additionally adapt the background from the input sample.

We note that a concurrent work produces realistic words~\cite{Kang20} using a
\ac{gan}-based system. This could partially be used in our method for improved
stylization, with the advantage that our method produces full line images.

\bibliographystyle{splncs04}
\bibliography{egbib}

\begin{thebibliography}{10}
\providecommand{\url}[1]{\texttt{#1}}
\providecommand{\urlprefix}{URL }
\providecommand{\doi}[1]{https://doi.org/#1}

\bibitem{Aksan19}
{Aksan}, E., {Hilliges}, O.: {STCN: Stochastic Temporal Convolutional
  Networks}. arXiv e-prints arXiv:1902.06568 (Feb 2019)

\bibitem{Aksam18}
Aksan, E., Pece, F., Hilliges, O.: Deepwriting: Making digital ink editable via
  deep generative modeling. In: Proceedings of the 2018 CHI Conference on Human
  Factors in Computing Systems. CHI ’18, Association for Computing Machinery,
  New York, NY, USA (2018)

\bibitem{Alonso19}
{Alonso}, E., {Moysset}, B., {Messina}, R.: Adversarial generation of
  handwritten text images conditioned on sequences. In: 2019 International
  Conference on Document Analysis and Recognition (ICDAR). pp. 481--486 (Sep
  2019)

\bibitem{Bai20}
{Bai}, X., {Ye}, L., {Zhu}, J., {Zhu}, L., {Komura}, T.: Skeleton filter: A
  self-symmetric filter for skeletonization in noisy text images. IEEE
  Transactions on Image Processing  \textbf{29},  1815--1826 (2020)

\bibitem{Balreira17}
{Balreira}, D.G., {Walter}, M.: Handwriting synthesis from public fonts. In:
  2017 30th SIBGRAPI Conference on Graphics, Patterns and Images (SIBGRAPI).
  pp. 246--253 (Oct 2017)

\bibitem{Chung15}
Chung, J., Kastner, K., Dinh, L., Goel, K., Courville, A.C., Bengio, Y.: A
  recurrent latent variable model for sequential data. In: Cortes, C.,
  Lawrence, N.D., Lee, D.D., Sugiyama, M., Garnett, R. (eds.) Advances in
  Neural Information Processing Systems 28, pp. 2980--2988. Curran Associates,
  Inc. (2015)

\bibitem{dijkstra59}
Dijkstra, E.W.: A note on two problems in connexion with graphs. Numerische
  mathematik  \textbf{1}(1),  269--271 (1959)

\bibitem{Goldhahn12}
Goldhahn, D., Eckart, T., Quasthoff, U.: Building large monolingual
  dictionaries at the leipzig corpora collection: From 100 to 200 languages.
  pp. 759--765 (2012)

\bibitem{Gomez19}
{Gomez}, R., {Furkan Biten}, A., {Gomez}, L., {Gibert}, J., {Karatzas}, D.,
  {Rusiñol}, M.: Selective style transfer for text. In: 2019 International
  Conference on Document Analysis and Recognition (ICDAR). pp. 805--812 (Sep
  2019)

\bibitem{graves}
{Graves}, A.: {Generating Sequences With Recurrent Neural Networks}. arXiv
  e-prints arXiv:1308.0850 (Aug 2013)

\bibitem{Haines2016}
Haines, T.S.F., Mac~Aodha, O., Brostow, G.J.: My text in your handwriting. ACM
  Trans. Graph.  \textbf{35}(3) (May 2016)

\bibitem{Isola17}
{Isola}, P., {Zhu}, J., {Zhou}, T., {Efros}, A.A.: Image-to-image translation
  with conditional adversarial networks. In: 2017 IEEE Conference on Computer
  Vision and Pattern Recognition (CVPR). pp. 5967--5976 (July 2017)

\bibitem{Kang20}
{Kang}, L., {Riba}, P., {Wang}, Y., {Rusi{\~n}ol}, M., {Forn{\'e}s}, A.,
  {Villegas}, M.: {GANwriting: Content-Conditioned Generation of Styled
  Handwritten Word Images}. arXiv e-prints arXiv:2003.02567 (Mar 2020)

\bibitem{cvl}
Kleber, F., Fiel, S., Diem, M., Sablatnig, R.: {CVL-DataBase: An Off-Line
  Database for Writer Retrieval, Writer Identification and Word Spotting}. In:
  Proceedings of the 2013 12th International Conference on Document Analysis
  and Recognition. pp. 560--564. ICDAR '13, IEEE Computer Society, Washington,
  DC, USA (2013)

\bibitem{Lee94}
Lee, T., Kashyap, R., Chu, C.: Building skeleton models via 3-d medial surface
  axis thinning algorithms. CVGIP: Graphical Models and Image Processing
  \textbf{56}(6),  462 -- 478 (1994)

\bibitem{Lian18}
Lian, Z., Zhao, B., Chen, X., Xiao, J.: Easyfont: A style learning-based system
  to easily build your large-scale handwriting fonts. ACM Trans. Graph.
  \textbf{38}(1) (Dec 2018)

\bibitem{iam-online}
Liwicki, M., Bunke, H.: {IAM-OnDB - an On-Line English Sentence Database
  Acquired from Handwritten Text on a Whiteboard}. In: 8th Intl. Conf. on
  Document Analysis and Recognition. vol.~2, pp. 956--961 (2005)

\bibitem{Nakamura18}
{Nakamura}, K., {Miyazaki}, E., {Nitta}, N., {Babaguchi}, N.: Generating
  handwritten character clones from an incomplete seed character set using
  collaborative filtering. In: 2018 16th International Conference on Frontiers
  in Handwriting Recognition (ICFHR). pp. 68--73 (Aug 2018)

\bibitem{nicolaou2015sparse}
Nicolaou, A., Bagdanov, A.D., Liwicki, M., Karatzas, D.: Sparse radial sampling
  lbp for writer identification. In: 2015 13th International Conference on
  Document Analysis and Recognition (ICDAR). pp. 716--720. IEEE (2015)

\bibitem{perez2003poisson}
P{\'e}rez, P., Gangnet, M., Blake, A.: Poisson image editing. In: ACM SIGGRAPH
  2003 Papers, pp. 313--318 (2003)

\bibitem{Ronneberger15}
Ronneberger, O., Fischer, P., Brox, T.: U-net: Convolutional networks for
  biomedical image segmentation. In: Navab, N., Hornegger, J., Wells, W.M.,
  Frangi, A.F. (eds.) Medical Image Computing and Computer-Assisted
  Intervention -- MICCAI 2015. pp. 234--241. Springer International Publishing,
  Cham (2015)

\bibitem{seuret2015gradient}
Seuret, M., Chen, K., Eichenbergery, N., Liwicki, M., Ingold, R.:
  Gradient-domain degradations for improving historical documents images layout
  analysis. In: 2015 13th International Conference on Document Analysis and
  Recognition (ICDAR). pp. 1006--1010. IEEE (2015)

\bibitem{Shen17}
{Shen}, W., {Zhao}, K., {Jiang}, Y., {Wang}, Y., {Bai}, X., {Yuille}, A.:
  Deepskeleton: Learning multi-task scale-associated deep side outputs for
  object skeleton extraction in natural images. IEEE Transactions on Image
  Processing  \textbf{26}(11),  5298--5311 (Nov 2017)

\bibitem{Yang19}
{Yang}, S., {Liu}, J., {Yang}, W., {Guo}, Z.: Context-aware text-based binary
  image stylization and synthesis. IEEE Transactions on Image Processing
  \textbf{28}(2),  952--964 (Feb 2019)

\bibitem{Zhang84}
Zhang, T.Y., Suen, C.Y.: {A Fast Parallel Algorithm for Thinning Digital
  Patterns}. Commun. ACM  \textbf{27}(3),  236--239 (Mar 1984)

\bibitem{Zhu17}
{Zhu}, J., {Park}, T., {Isola}, P., {Efros}, A.A.: Unpaired image-to-image
  translation using cycle-consistent adversarial networks. In: 2017 IEEE
  International Conference on Computer Vision (ICCV). pp. 2242--2251 (Oct 2017)

\end{thebibliography}


\begin{thebibliography}{1}
\providecommand{\url}[1]{\texttt{#1}}
\providecommand{\urlprefix}{URL }
\providecommand{\doi}[1]{https://doi.org/#1}

\bibitem{Christlein18DAS}
Christlein, V., Maier, A.: {Encoding CNN Activations for Writer Recognition}.
  In: 13th IAPR International Workshop on Document Analysis Systems. pp.
  169----174 (2018)

\bibitem{Jegou12ALI}
J{\'{e}}gou, H., Perronnin, F., Douze, M., S{\'{a}}nchez, J., P{\'{e}}rez, P.,
  Schmid, C.: {Aggregating Local Image Descriptors into Compact Codes}. Pattern
  Analysis and Machine Intelligence, IEEE Transactions on  \textbf{34}(9),
  1704--1716 (2012)

\bibitem{Lowe04}
Lowe, D.G.: {Distinctive Image Features from Scale-Invariant Keypoints}.
  International Journal of Computer Vision  \textbf{60}(2),  91--110 (2004)

\bibitem{Murray16}
Murray, N., Jegou, H., Perronnin, F., Zisserman, A.: {Interferences in Match
  Kernels}. IEEE Transactions on Pattern Analysis and Machine Intelligence
  \textbf{39}(9),  1797--1810 (2016)

\bibitem{nicolaou2015sparse}
Nicolaou, A., Bagdanov, A.D., Liwicki, M., Karatzas, D.: Sparse radial sampling
  lbp for writer identification. In: 2015 13th International Conference on
  Document Analysis and Recognition (ICDAR). pp. 716--720. IEEE (2015)

\end{thebibliography}

\end{document}


\title{Spatio-Temporal Handwriting Imitation}

\titlerunning{Spatio-Temporal Handwriting Imitation}
\author{Martin Mayr\orcidID{0000-0002-3706-285X} \and
Martin Stumpf\orcidID{0000-0002-2278-8500} \and \\
Anguelos Nicolaou\orcidID{0000-0003-3818-8718} \and
Mathias Seuret\orcidID{0000-0001-9153-1031} \and
Andreas Maier\orcidID{0000-0002-9550-5284} \and
Vincent Christlein\orcidID{0000-0003-0455-3799}}
\authorrunning{M.\ Mayr et al.}
\institute{Pattern Recognition Lab, Friedrich–Alexander-Universität Erlangen–Nürnberg
\email{\{firstname.lastname\}@fau.de}\\
\url{https://lme.tf.fau.de/}}

\title{Supplementary Material to ``Spatio-Temporal Handwriting Imitation''}
\titlerunning{Spatio-Temporal Handwriting Imitation (Supplementary Material)}

\maketitle

\renewcommand\thesection{\Alph{section}}

\section{Additional Experiments}
\subsection{Learning-free WI Algorithm}
\Cref{tab:simple_wi} shows all combinations of query and retrieval database between different modalities performed with the learning-free writer identification algorithm \cite{nicolaou2015sparse}.
All experiments operate on text paragraphs from 27 writer identities as in the train-set of the CVL dataset.
Specifically we produced three modalities for every sample, one with the proposed pipeline, one with the proposed pipeline substituting the pix2pix skeleton with naive skeletons, and a real sample from the CVL train-set.
Every modality is represented by an in-vocabulary (IV) and an out-of-vocabulary (OV) sample.
Fake and real IV samples share the same transcription they are page 4 from the CVL dataset.
While OV samples fake samples share the same transcription among them and OV real samples are page 1 from the CVL dataset.
\emph{Soft$X$} refers to the usual TOP$X$ rates, \ie average percentage that the correct writers are among the first $X$ ranks.

\begin{table}
\centering
    \caption{Evaluation of WI  in various modalities of the data, metrics given in percent.}
    \label{tab:simple_wi}
    \begin{tabular}{llc@{\hspace{1em}}cccccc}
    \toprule

Query & DB & Skeleton & mAP & Acc. & Soft2 & Soft3 & Soft4 & Soft5 \\
\midrule

Fake (OV) & Real(IV) & pix2pix & 48.6   & 33.3   & 44.4   & 51.8   & 62.9   & 74.0    \\

Fake (OV) & Fake(IV) & pix2pix & 100.0   & 100.0   & 100.0   & 100.0   & 100.0   & 100.0    \\

Fake (OV) & Real(OV) & pix2pix & 49.2   & 33.3   & 44.4   & 62.9   & 66.6   & 70.3    \\

Fake (IV) &Real (IV) & pix2pix & 51.0   & 37.0   & 51.8   & 51.8   & 55.5   & 74.1    \\

Fake (IV) & Fake(OV) & pix2pix & 100.0   & 100.0   & 100.0   & 100.0   & 100.0   & 100.0    \\

Fake (IV) & Real(OV) & pix2pix & 52.3   & 40.7   & 44.4   & 59.2   & 62.9   & 66.6    \\

Fake (OV) & Real (IV) & naive & 54.5   & 40.7   & 51.8   & 62.9   & 70.3   & 74.0    \\

Fake (OV)& Fake(IV) & naive & 100.0   & 100.0   & 100.0   & 100.0   & 100.0   & 100.0    \\

Fake (OV) & Real(OV) & naive & 53.9   & 40.7   & 48.1   & 66.6   & 66.6   & 66.6    \\

Fake (IV) &Real (IV) & naive & 49.0   & 29.6   & 51.8   & 62.9   & 74.0   & 74.0    \\

Fake (IV)& Fake (OV) & naive & 100.0   & 100.0   & 100.0   & 100.0   & 100.0   & 100.0    \\

Fake (IV) & Real (OV) & naive & 52.1   & 37.0   & 51.8   & 62.9   & 66.6   & 66.6    \\

Fake (IV) & Fake(OV) & naive/pix2pix & 100.0   & 100.0   & 100.0   & 100.0   & 100.0   & 100.0    \\

Fake (IV) & Fake(OV) & naive & 100.0   & 100.0   & 100.0   & 100.0   & 100.0   & 100.0    \\

Real (IV) & Real(OV) & naive & 100.0   & 100.0   & 100.0   & 100.0   & 100.0   & 100.0    \\
    \bottomrule
    \end{tabular}
\end{table}

\subsection{Alternative WI Algorithm + Experiment using Full CVL Dataset}
\begin{table}[t]
    \caption{Evaluation of WI using a different WI algorithm and different Query/DB settings, metrics given in percent.
    \subref{tab:single} Human page 4 as database. 
    \subref{tab:multiple} All human pages as database.}
    \label{tab:sift_wi}
    \centering
    \begin{tabular}{lc@{\hspace{1em}}cc@{\hspace{1em}}cc}
    \toprule
    & & \multicolumn{2}{c}{(a) Single}\phantomsubcaption\label{tab:single} & \multicolumn{2}{c}{(b) Full\phantomsubcaption\label{tab:multiple}}\\
    \cmidrule(lr){3-4} \cmidrule(lr){5-6}
    Query  & Skeleton & mAP & Acc. & mAP & Acc. \\
    \midrule
    Fake (OV) & Naive   & 44.9 & 29.6 & 32.4 & 29.6 \\
    Fake (IV) & Naive   & 39.3 & 18.5 & 32.7 & 29.6 \\
    Fake (OV) & pix2pix & 46.0 & 29.6 & 38.5 & 40.7 \\
    Fake (IV) & pix2pix & 47.5 & 29.6 & 36.0 & 37.0 \\
    \bottomrule
    \end{tabular}
\end{table}

\Cref{tab:sift_wi} shows additional writer identification results, where we evaluated a different writer identification pipeline.
Therefore, we used a method by Christlein et al.~\cite{Christlein18DAS}, but replaced the CNN-based features with dirichlet-normalized SIFT~\cite{Lowe04} descriptors, which are whitened and dimensionality-reduced to 64 components by means of PCA. 
The rest of the pipeline stays unchanged, \ie VLAD encodings~\cite{Jegou12ALI} are computed using 100 clusters for $k$-means, which are aggregated using generalized max pooling~\cite{Murray16} ($\lambda=1000$).
This process is repeated three times. 
All resulting representations are then jointly whitened by another PCA.
For computing the $k$-means clusters and PCA matrices, we used the CVL test dataset. 
Similar to the learning-free method, which was employed throughout the main paper and in \cref{tab:simple_wi}, \cref{tab:single} shows that the method works similarly well whether it uses non-vocabulary words or not. Furthermore, the skeletonization we proposed using the adapted pix2pix mesh has improved the skeletonization. The same behavior can be observed in \cref{tab:multiple}.

\section{User Study Details}

\begin{figure}[t]
    \centering
    \includegraphics[width=\textwidth]{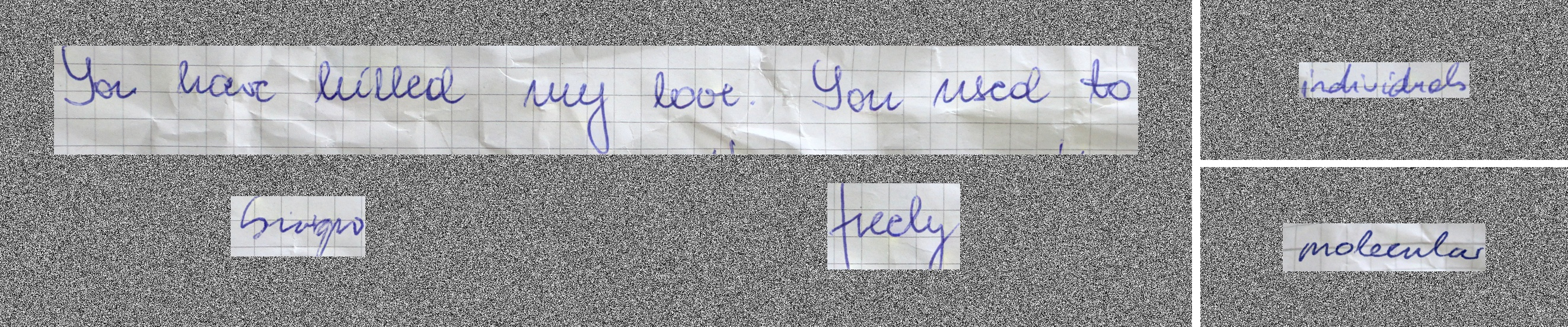}
    \caption{Contextualized (left) and non-contextualized samples (right) as seen in the user study.}
    \label{fig:user-study-example}
\end{figure}

In \cref{fig:user-study-example} a contextualized and two non-contextualized samples can be seen as they were presented to the subjects. The contextualized samples are used for evaluating the \style task. They consist of an anchor sentence, one synthetic word and one real word. The subject has to decide which word is written by the same person as the anchor sentence. 
The non-contextual samples are used for the \turing task. There only one sample is shown at a time and the subject has to decide if the shown sample is human-written or synthetic.

\begin{table}[t]
    \caption{Number of samples per category}
    \label{tab:numbers}
    \centering
    \begin{tabular}{l@{\hspace{1em}}cc@{\hspace{1em}}c}
    \toprule
    & IV & OV & Total\\
    \midrule
    \turing & 10 & 6 & 16\\
    \style  & 29 & 35 & 64\\
         \bottomrule
    \end{tabular}
\end{table}

\Cref{tab:numbers} shows the number of samples of in-vocabulary words (IV) and out-of-vocabulary words (OV) for both tasks.
Note that for all of the shown samples in \cref{fig:user-study-example} an artificial background was added. In \SI{50}{\percent} of the queries the background was changed from white to notebook paper. 
The study with all samples can be viewed under this link \url{https://forms.gle/MGCPk5UkxnR23FqT9}.

\section{Additional Analyses}

\subsection{Qualitative Results -- Paragraphs}
\begin{figure}[h!]
    \centering
    \begin{tabularx}{\textwidth}{YY}
         \toprule
         \includegraphics[width=0.4\textwidth]{img/top_flop/sentence/A_0012-07-00.png} &  
         \includegraphics[width=0.4\textwidth]{img/top_flop/sentence/A_0022-07-00.png}
         \\ \midrule
         \includegraphics[height=0.15\textheight]{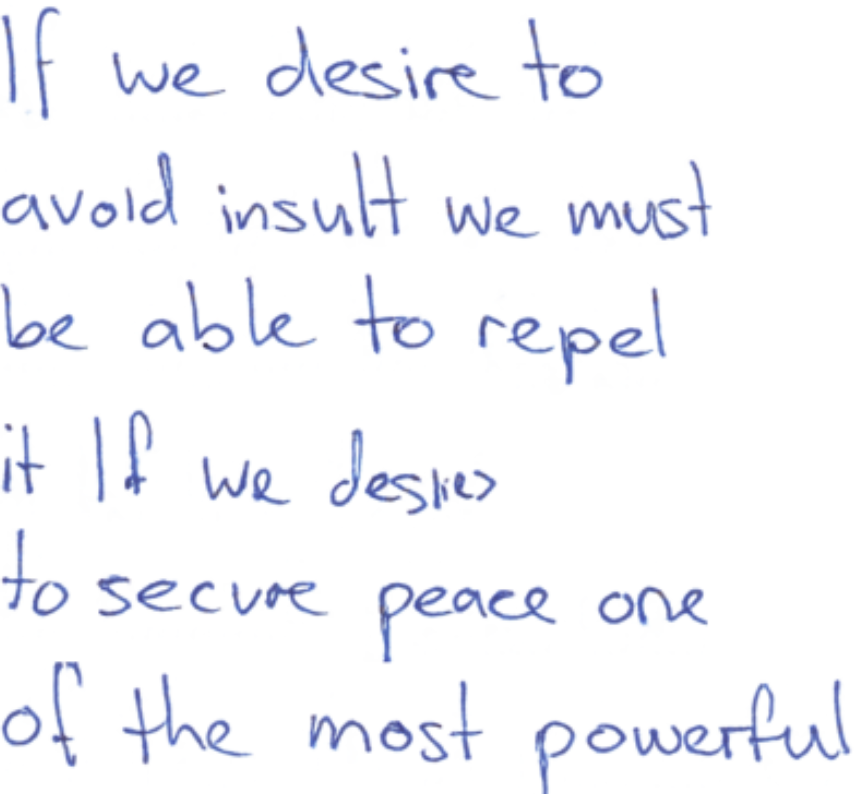} &
         \includegraphics[height=0.15\textheight]{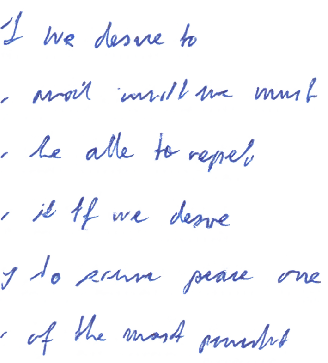}
         \\ \bottomrule
    \end{tabularx}
    \caption{Style input lines of two writers with their corresponding generated paragraphs.
   } 
    \label{tab:example_paragraphs}
\end{figure}
\cref{tab:example_paragraphs} illustrates different outputs of new paragraphs, \ie containing out-of-vocabulary words.
The left generated text looks convincing and is legible. Conversely, the right sample shows the output of a writer with a rather bad handwriting. 
This leads to artifacts and an illegible output.
Due to the fact that there is no ground truth information of the punctuation, the pipeline sometimes produces artifacts at the beginning of the line when the style contains a punctuation.

\subsection{User Study Results}

\begin{figure}[t]
	\centering
 	\input{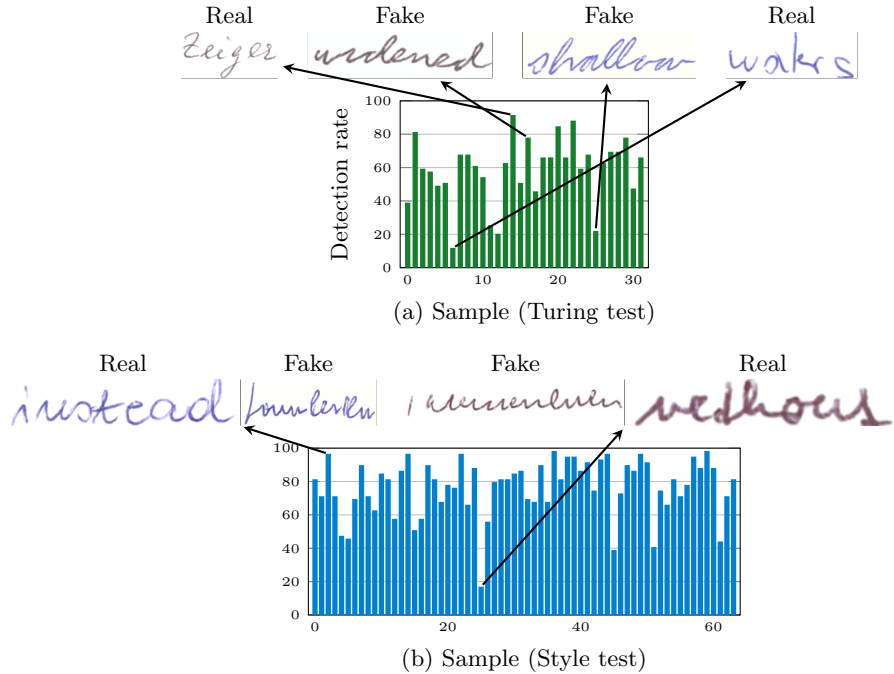} 
	\caption{%
		Per-sample results. Additionally, showing best and worst cases for real and
		fake samples, respectively.
	}
  \label{fig:per_sample}
\end{figure}
\Cref{fig:per_sample} shows the mean accuracy for each 96 samples. Additionally,
the samples that were detected the best and the worst are depicted.
It is clearly visible that the detection results of the subjects highly increased when an anchor sentence of the writer was given. 

\begin{figure}[t]
 	\begin{tikzpicture}
\begin{groupplot}[%
	group style={group size=4 by 1, horizontal sep=0.7cm},
	width=0.318\textwidth,
	height=3.8cm,
	ymin=0, ymax=100,
	ylabel near ticks,
	xlabel near ticks,
	label style={font=\footnotesize},
	ymajorgrids=true,
	major x tick style=transparent,
	tick label style={font=\tiny},
		]
	\nextgroupplot[%
	ybar, %
	xlabel={(a) H\phantomsubcaption\label{fig:turing_h}},
	xticklabel style={align=center,yshift=0.5em},
	ylabel={Detection rate},
	bar width=1pt,
	]

	\addplot+[fill=uncontext,draw opacity=0] table 
		[x expr=\coordindex,y=un] {figures/h.txt};

	\nextgroupplot[%
	xlabel={(b) CV\phantomsubcaption\label{fig:turing_cv}},
	ybar, %
	xticklabel style={align=center,yshift=0.5em},
	bar width=1pt,
	]

	\addplot+[fill=uncontext,draw opacity=0] table 
		[x expr=\coordindex,y=un] {figures/cv.txt};

	\nextgroupplot[%
	xlabel={(c) CS\phantomsubcaption\label{fig:turing_cs}},
	ybar, %
	xticklabel style={align=center,yshift=0.5em},
	bar width=1pt,
	]

	\addplot+[fill=uncontext,draw opacity=0] table 
		[x expr=\coordindex,y=un] {figures/cs.txt};
	\nextgroupplot[%
	xlabel={(d) O\phantomsubcaption\label{fig:turing_o}},
	ybar, %
	xticklabel style={align=center,yshift=0.5em},
	bar width=1pt,
	]

	\addplot+[fill=uncontext,draw opacity=0] table 
		[x expr=\coordindex,y=un] {figures/o.txt};

\end{groupplot}
\end{tikzpicture} 
  \caption{All individual results (\turing test)}
  \label{fig:individual_turing}
\end{figure}
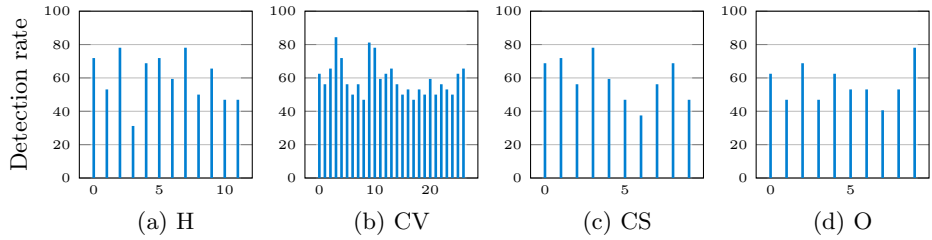

\begin{figure}[t]
 	\begin{tikzpicture}
\begin{groupplot}[%
	group style={group size=4 by 1, horizontal sep=0.7cm},
	width=0.318\textwidth,
	height=3.8cm,
	ymin=0, ymax=100,
	ylabel near ticks,
	xlabel near ticks,
	label style={font=\footnotesize},
	ymajorgrids=true,
	major x tick style=transparent,
	tick label style={font=\tiny},
		]
	\nextgroupplot[%
	ybar, %
	xlabel={(a) H\phantomsubcaption\label{fig:style_h}},
	xticklabel style={align=center,yshift=0.5em},
	ylabel={Detection rate},
	bar width=1pt,
	]

	\addplot+[fill=context,draw opacity=0] table 
		[x expr=\coordindex,y=con] {figures/h.txt};

	\nextgroupplot[%
	xlabel={(b) CV\phantomsubcaption\label{fig:style_cv}},
	ybar, %
	xticklabel style={align=center,yshift=0.5em},
	bar width=1pt,
	]

	\addplot+[fill=context,draw opacity=0] table 
		[x expr=\coordindex,y=con] {figures/cv.txt};

	\nextgroupplot[%
	xlabel={(c) CS\phantomsubcaption\label{fig:style_cs}},
	ybar, %
	xticklabel style={align=center,yshift=0.5em},
	bar width=1pt,
	]

	\addplot+[fill=context,draw opacity=0] table 
		[x expr=\coordindex,y=con] {figures/cs.txt};
	\nextgroupplot[%
	xlabel={(d) O\phantomsubcaption\label{fig:style_o}},
	ybar, %
	xticklabel style={align=center,yshift=0.5em},
	bar width=1pt,
	]

	\addplot+[fill=context,draw opacity=0] table 
		[x expr=\coordindex,y=con] {figures/o.txt};

\end{groupplot}
\end{tikzpicture} 
  \caption{All individual results (\style test)}
  \label{fig:individual_style}
\end{figure}
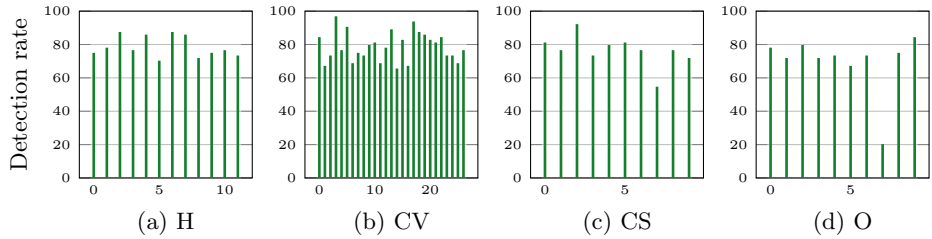
\Cref{fig:individual_turing} and \cref{fig:individual_style} show the
individual results of all participants for the \turing and the \style tests. 
The user study comprised 59 individuals, 12 humanities scholars (H),
27 people of computer vision (CV), 10 computer scientists (CS), 10 others (O).

\section{Failure Cases}

During our evaluation, we observed errors which occur more frequently in different stages of the pipeline.
Note that these artifacts are often not severe and through the last pen style transfer, most of these artifacts vanish again. 
However, they show potential improvements for future work. 

\paragraph{Skeletonization}

\begin{figure}
    \centering
    \subcaptionbox{\label{fig:primitive_in} real input image}{
    \fbox{\includegraphics[width=0.25\textwidth]{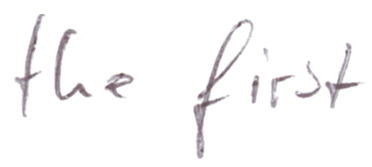}
    }}
    \subcaptionbox{\label{fig:primitive_out} primitive skeletonization}{
    \fbox{\includegraphics[width=0.25\textwidth]{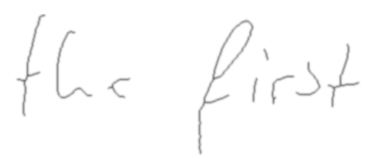}
    }}
    \caption{Typical failure modes of the primitive skeletonization. 
    }
    \label{fig:primitive_skeletonization_failures}

\vspace{0.5cm}
    \centering
    \subcaptionbox{\label{fig:skeleton_failure_0}}{
    \fbox{\includegraphics[height=0.14\textheight]{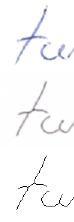}
    }}
    \subcaptionbox{\label{fig:skeleton_failure_1}}{
    \fbox{\includegraphics[height=0.14\textheight]{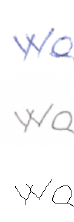}
    }}
    \subcaptionbox{\label{fig:skeleton_failure_2}}{
    \fbox{\includegraphics[height=0.14\textheight]{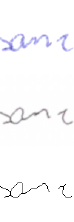}
    }}
    \subcaptionbox{\label{fig:skeleton_failure_3}}{
    \fbox{\includegraphics[height=0.14\textheight]{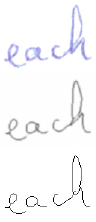}
    }}
    \subcaptionbox{\label{fig:skeleton_failure_4}}{
    \fbox{\includegraphics[height=0.14\textheight]{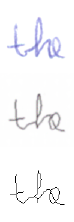}
    }}
    \caption{Typical failure modes of the final skeletonization network.}
    \label{fig:learned_skeletonization_failures}
\vspace{0.5cm}

    \centering
    \subcaptionbox{\label{fig:online_failure_0}}{
    \fbox{\includegraphics[height=0.08\textheight]{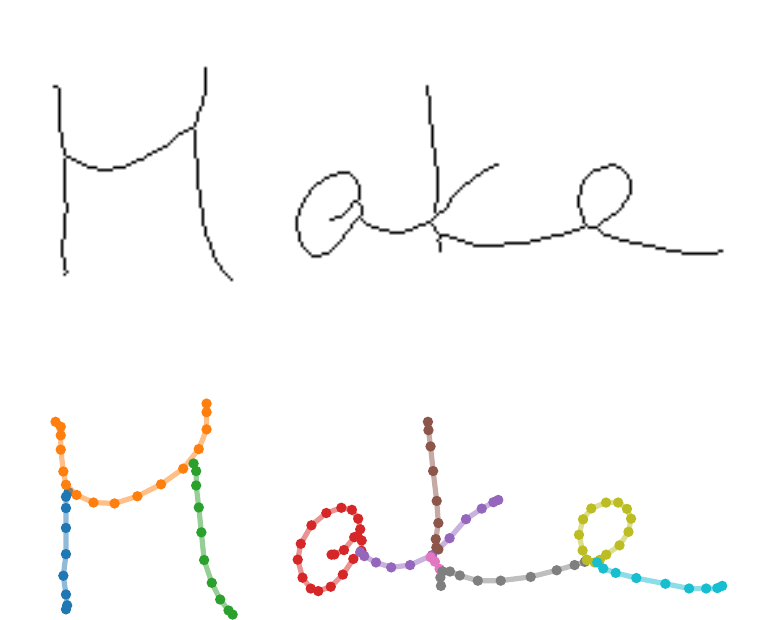}
    }}
  \subcaptionbox{\label{fig:online_failure_1}}{
    \fbox{\includegraphics[height=0.08\textheight]{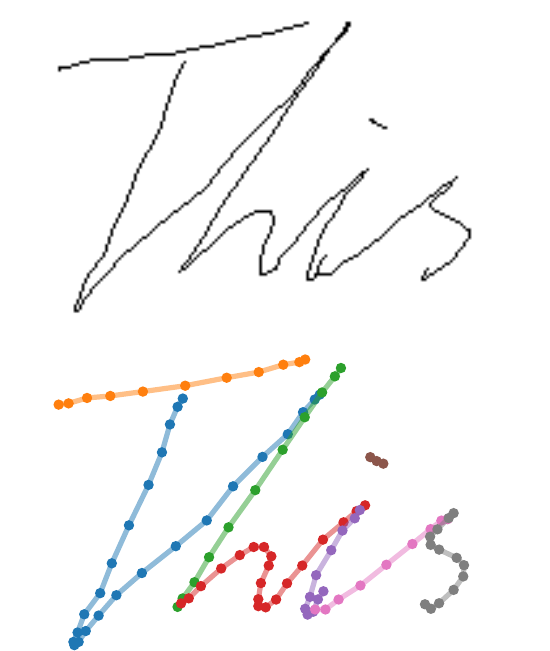}
    }}
  \subcaptionbox{\label{fig:online_failure_2}}{
    \fbox{\includegraphics[height=0.08\textheight]{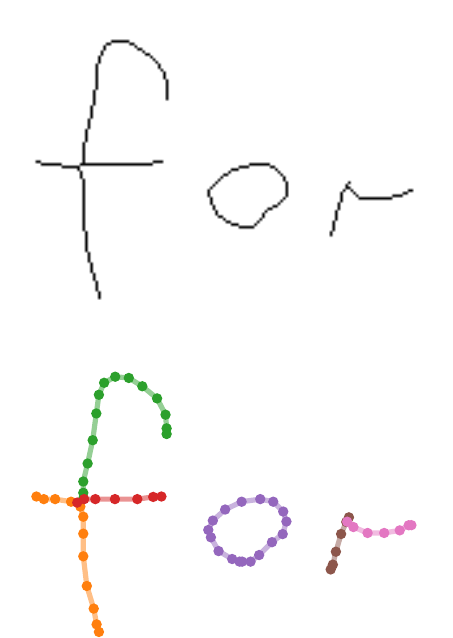}
    }}
    \caption{Typical failure modes of the online approximation. }
    \label{fig:online_errors}

\vspace{0.5cm}

  \centering
  \subcaptionbox{\label{fig:WST_failure_0}}{
    \fbox{\includegraphics[height=0.03\textheight]{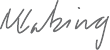}
    }}
  \subcaptionbox{\label{fig:WST_failure_1}}{
    \fbox{\includegraphics[height=0.03\textheight]{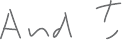}
    }}
  \subcaptionbox{\label{fig:WST_failure_2}}{
    \fbox{\includegraphics[height=0.03\textheight]{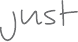}
    }}
  \subcaptionbox{\label{fig:WST_failure_3}}{
    \fbox{\includegraphics[height=0.03\textheight]{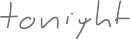}
    }}
  \caption{Typical failure modes of the writer style transfer.}
  \label{fig:writerStyleTransferFailures}
\end{figure}

The results of the \textit{primitive skeletonization} often show merging of adjacent lines, breaking apart of continuous lines at intersections, removal of details or cutting corners, as seen in \cref{fig:primitive_skeletonization_failures}.

Also, \textit{learned skeletonization} produces some common failure cases, described in \cref{fig:learned_skeletonization_failures}. From top to bottom: input image, skeletonization network output, final skeleton.
\cref{fig:skeleton_failure_0} and \cref{fig:skeleton_failure_1} show that crossing lines tend to be broken apart. \cref{fig:skeleton_failure_2} shows the loss of detail by over-smoothing. \cref{fig:skeleton_failure_3} shows another example where crossing lines get broken apart at the \emph{e} and merged lines at the \emph{h}, caused by the deblurring step. \cref{fig:skeleton_failure_4} shows a combination of them. The \emph{h} gets distorted by both the network and the deblurring step, and the \emph{e} ends up touching the \emph{h}.
The comparison of both methods is outlined in the main text of the paper.

\paragraph{Approximation of Online Representation}

\cref{fig:online_errors} shows the most common errors occurring during the approximation of the online representation. 
The \emph{M} of \cref{fig:online_failure_0} demonstrates the inability of the algorithm to understand lines that are drawn on top of each other, causing the continuous line to be split into three segments. Further, the \emph{k} and the \emph{e} show cases where the algorithm fails to connect crossing lines properly. \cref{fig:online_failure_1} describes the problem that the algorithm has with sharp edges. The entire word is mostly one continuous line, but the algorithm split it into lots of segments. \cref{fig:online_failure_2} is an example for intersecting lines of the \emph{f} which got connected incorrectly.

\paragraph{Writer Style Transfer}

The stage of the writer style transfer produces some common failures, like addition of superfluous lines (\cref{fig:WST_failure_0}), incorrect splitting of lines (\cref{fig:WST_failure_1}),
incorrect positioning of lines,
detail removal, like \emph{t}-, \emph{f}-lines or \emph{i}-dots (\cref{fig:WST_failure_2}),
skipping letters at the start of the message, which is an artifact that happens when the network was unable to correctly parse the input style.

\clearpage
\bibliographystyle{splncs04}
\bibliography{egbib}